\documentclass[10pt,twocolumn,letterpaper,dvipsnames]{article}
\usepackage[pagenumbers]{cvpr}  

\usepackage{amsmath,amssymb,amsfonts}  
\usepackage{bm}
\usepackage{csquotes}                  
\usepackage{etoolbox}                  
\usepackage{graphicx}
\usepackage{nicefrac}                  
\usepackage{soul}                      
\usepackage{tikz}                      
\usepackage{textcomp}                  
\usepackage{url}                       
\usepackage{xcolor}                    
\usepackage{xparse}                    

\definecolor{cvprblue}{rgb}{0.21,0.49,0.74}
\usepackage[pagebackref,breaklinks,colorlinks,citecolor=cvprblue]{hyperref}
\usepackage[capitalize]{cleveref}      
\usepackage[accsupp]{axessibility}     

\newtoggle{nographics}\togglefalse{nographics}

\newtoggle{nocomments}\togglefalse{nocomments}

\begin{document}
    \title{NeuSD: Surface Completion with Multi-View Text-to-Image Diffusion}

\author{\
\makebox[\textwidth][c]{
\hfill{}Savva Ignatyev\textsuperscript{1*}
\hfill{}Daniil Selikhanovych\textsuperscript{1*}
\hfill{}Oleg Voynov\textsuperscript{1,2}
\hfill{}Yiqun Wang\textsuperscript{3}
\hfill{}}\\
\makebox[\textwidth][c]{
\hfill{}Peter Wonka\textsuperscript{4}
\hfill{}Stamatios Lefkimmiatis\textsuperscript{5}
\hfill{}Evgeny Burnaev\textsuperscript{1,2}
\hfill{}}
\medskip\\\
\makebox[\textwidth][c]{
\hfill{}\textsuperscript{1}Skoltech, Russia
\hfill{}\textsuperscript{2}AIRI, Russia
\hfill{}\textsuperscript{3}Chongqing University, China
\hfill{}}\\
\makebox[\textwidth][c]{
\hfill{}\textsuperscript{4}KAUST, Saudi Arabia
\hfill{}\textsuperscript{5}AI Foundation and Algorithm Lab, Russia
\hfill{}}
}

    \newcommand{\todo}[1]{{\color{red}TODO: #1}}

\DeclareGraphicsExtensions{.eps,.pdf,.jpg,.png}
\graphicspath{{src/img/}}
\makeatletter
\iftoggle{nographics}{
    \LetLtxMacro{\includegraphics@orig}{\includegraphics}
    \RenewDocumentCommand{\includegraphics}{ s O{} m }{%
            {\setlength{\fboxsep}{0pt}%
        \colorbox{lightgray}{\phantom{\IfBooleanTF{#1}{\includegraphics@orig*}{\includegraphics@orig}[#2]{#3}}}%
        }%
    }
}{}
\makeatother

    \twocolumn[{%
    \renewcommand\twocolumn[1][]{#1}%
    \maketitle
    \vspace{-10mm}
    \begin{center}
        \begin{tikzpicture}
            \node[anchor=south west,inner sep=0] (image) at (0,0) {\includegraphics[width=\textwidth]{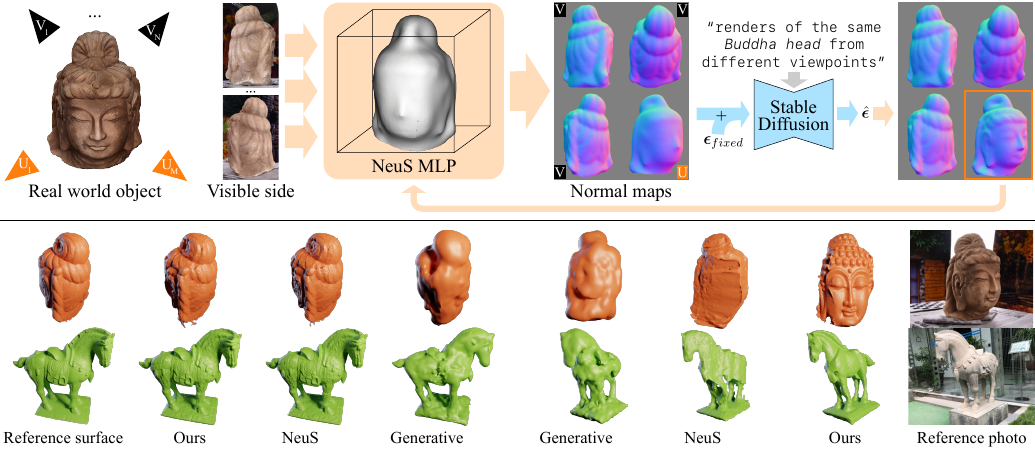}};
            \begin{scope}[shift=(image.south west), x=1pt, y=1pt, anchor=south west, inner sep=0, font=\small]
                \node at (148.5, 1.3) {\phantom{S}~\cite{wang2021neus}};
                \node at (219, 1.3) {\phantom{e}~\cite{tang2023dreamgaussian}};
                \node at (290.5, 1.3) {\phantom{e}~\cite{tang2023dreamgaussian}};
                \node at (341.5, 1.3) {\phantom{S}~\cite{wang2021neus}};
            \end{scope}
        \end{tikzpicture}
        \captionof{figure}{
            We consider reconstruction of a partially observed surface.
            We train a neural implicit surface representation using the photos of the observed part while guiding the training using a 2D diffusion model.
            Our key insights are applying the diffusion model to normal maps instead of color renderings (\cref{sec:normal_sds}),
            freezing the noise \(\bm{\epsilon}_{fixed}\) added to the input of the model (\cref{sec:frozen_sds}),
            and applying the model to a grid of normal maps obtained from visible (V) and unobserved (U) sides of the surface (\cref{sec:mv_sds}).
            Our method accurately reconstructs the surface from the observed side (bottom left)
            and completes it from the unobserved side in a plausible manner (bottom right).
        }
        \label{fig:fig_teaser}
    \end{center}
}]

\def\thefootnote{*}\footnotetext{Equal contribution}
\def\thefootnote{}\footnotetext{For correspondence: o.voynov@skoltech.ru}

\begin{abstract}
    We present a novel method for 3D surface reconstruction from multiple images where only a part of the object of interest is captured.
    Our approach builds on two recent developments:
    surface reconstruction using neural radiance fields for the reconstruction of the visible parts of the surface,
    and guidance of pre-trained 2D diffusion models in the form of Score Distillation Sampling (SDS) to complete the shape in unobserved regions in a plausible manner.
    We introduce three components.
    First, we suggest employing normal maps as a pure geometric representation for SDS instead of color renderings which are entangled with the appearance information.
    Second, we introduce the freezing of the SDS noise during training which results in more coherent gradients and better convergence.
    Third, we propose Multi-View SDS as a way to condition the generation of the non-observable part of the surface without fine-tuning or making changes to the underlying 2D Stable Diffusion model.
    We evaluate our approach on the BlendedMVS dataset demonstrating significant qualitative and quantitative improvements over competing methods.
\end{abstract}

\section{Introduction}
\label{sec:intro}
We propose a novel framework called \emph{NeuSD} that addresses the problem of 3D surface reconstruction from a set of the posed RGB input images. Specifically, we consider the case where the space of camera locations of the input images around an object is not evenly sampled. In practice, there are many realistic scenarios where a human observer can nicely capture images from a set of viewpoints from one side of the object, but not from the other side.
For example, imagine a statue in a museum placed right next to a wall or behind a fence or a car being parked close to a wall. This important problem has not been systematically studied in the literature and we set out to propose a solution to this challenge. We will first put this problem in the context of the existing literature.

The current state of the art in surface reconstruction uses neural implicit surfaces. Notably, \cite{wang2021neus} introduced NeuS, which
represents a surface as a signed distance field.
NeuS demonstrated reconstruction results on par with the classical Multi-View Stereo (MVS) methods, such as~\cite{schoenberger2016structure,schoenberger2016pixelwise},
and more importantly, outputs watertight meshes that are ready to be used in other vision or graphics applications.

Both traditional and deep learning approaches to 3D reconstruction usually operate under the assumption of full observability. Nevertheless, there are two strands of literature that are related to the problem of partially-observed input shapes.
One is shape completion. However, there are two reasons why this is typically viewed as a separate problem and solutions cannot easily be transferred to our problem setting. First, the majority of state-of-the-art approaches
are trained on category-based datasets like ShapeNet~\cite{chang2015shapenet},
which limits their applicability. Second, these methods use point clouds as input instead of a set of images. We, therefore, do not consider shape completion methods as direct competitors.
The other important line of work started with the seminal paper DreamFusion by~\cite{poole2022dreamfusion}. They suggested a new mechanism, Score Distillation Sampling (SDS),
which allowed us to employ pre-trained 2D text-to-image diffusion models as \textquote{critics} that provide the gradient to learn a 3D neural field. 
Since then, a substantial amount of work on this topic emerged to advance generative methods for 3D shape synthesis working either with a text prompt~\cite{zhu2023hifa} or a single input image (and a text prompt)~\cite{melas2023realfusion,liu2023one}.
We extend these excellent methods with two important new capabilities.
First, existing methods are not specialized in surface reconstruction and they use a volumetric instead of a surface representation to characterize the resulting 3D shape.
We provide a solution that can output clean surfaces in the form of distance fields.
Second, the single and few-shot methods do not generalize well to additional viewpoints and are often inherently limited in the viewpoints they can consider, requiring the view-dependent prompt or fixed w.r.t. each other camera positions.
By contrast, we propose a method that generalizes better to many input images from relatively closeby viewpoints, and we demonstrate this capability in our experimental validation.

Our proposed solution starts from NeuS as our baseline and complements it with SDS guidance using the Stable Diffusion model~\cite{rombach2022high}.
We further develop our method with the following three contributions:
(1) SDS Surface Shape Guidance: instead of using the RGB rendering for SDS guidance we use the normal map, which contains only the geometric information disentangled from the appearance;
(2) Frozen SDS: for faster convergence and improved quality, we fix the noise map used in SDS;
(3) Multi-View SDS: to condition the diffusion model on the visible part of the object we concatenate multiple normal maps in a grid for a joint SDS evaluation.

We test our approach on the BlendedMVS dataset~\cite{yao2020blendedmvs} and show that our method is preferred by CLIP scores and a user study (users preferred our method $>85\%$) compared to generative and reconstructive baselines.

\section{Related work}
\label{sec:related}

\noindent \textbf{Implicit surface reconstruction.}
Since its introduction, Neural Radiance Fields (NeRF)~\cite{mildenhall2021nerf} achieved impressive photorealistic results in novel view synthesis.
However, NeRF is a poor representation for multi-view surface reconstruction, since its density field is highly ambiguous and ill-regularized.
To adapt NeRF to surface reconstruction, several works~\cite{yariv2020multiview, yariv2021volume,wang2021neus} proposed Neural Implicit Surfaces.
NeuS~\cite{wang2021neus} specifically, implicitly parameterizes the NeRF density as a rapidly decreasing function of the signed distance to the surface (SDF),
and maintains the consistency of the SDF using an Eikonal penalty~\cite{gropp2020implicit}.
Subsequent works proposed constraining the implicit surface using the points obtained with Structure-from-Motion~\cite{fu2022geo},
improving multi-view consistency via a warping-based loss term~\cite{darmon2022improving},
improving performance in featureless areas via a normal-based regularization~\cite{wang2022neuris, wang2022neuralroom},
or improving accuracy and convergence speed by using multi-resolution hash encoding~\cite{wang2023neus2} or hybrid representations of the surface~\cite{dogaru2023sphere}.
While some works extended NeuS or NeRF to sparse image sampling~\cite{long2022sparseneus,yu2020pixelnerf},
still the vast majority of the work is concerned with the reconstruction of the visible part of the surface only,
and is not applicable as is in our setting with a partially observed surface.

\noindent \textbf{2D diffusion for 3D generation.}
Diffusion generative networks started as a fascinating concept and gradually made their way to the state of art through the recent years.
Their idea for inference is to start from the Gaussian noise and gradually remove noise with the help of the \emph{denoising network} step-by-step in the so-called \emph{reverse diffusion process}.
The training samples for the denoising network are drawn from the \emph{forward diffusion process}, where real photos are taken and the noise is added to them.
The denoising network can be conditioned on additional input such as 2D segmentation mask~\cite{zhang2023adding}, depth map, or text prompt.
Latent diffusion~\cite{rombach2022high} was the first work that achieved generating very high-quality 2D images from a general-case text prompt-conditioned generative image model.
Later, Stable Diffusion extended this work and went viral by producing near-photorealistic samples from user requests.

Despite its undeniable success and some signs that Stable Diffusion (SD) operates the internal 3D-like representation~\cite{chen2023beyond}, obtaining the 3D model from SD is a non-trivial task.
DreamFusion~\cite{poole2022dreamfusion} introduced the Score Distillation Sampling (SDS) approach, which allows propagation of the learning signal from a pre-trained 2D diffusion model into the 3D implicit neural representation model.
Score Jacobian Chaining~\cite{wang2023score} provided a theoretical framework for practically the same algorithm.
Multiple works have further improved the text-conditioned 3D generation.
Magic 3D~\cite{lin2023magic3d} introduced a two-stage approach, first optimizing the low-resolution radiance field, and then switching to high-resolution mesh-texture representation.
HIFA~\cite{zhu2023hifa} suggested gradually decreasing SDS noise and making multiple consecutive denoising steps before the backpropagation.

Recently, several works adapted 2D diffusion models for 3D generation conditioned on a single image, \ie, generative single-view 3D reconstruction.
RealFusion~\cite{melas2023realfusion} utilized textual inversion~\cite{gal2022image} to condition a pre-trained text-to-image diffusion model on the input image,
and then train a NeRF using SDS.
One-2-3-45~\cite{liu2023one} used a special multi-view diffusion model conditioned on the input image~\cite{Liu_2023_ICCV}
to synthesize images from different viewpoints and then train a neural surface representation on them.
DreamGaussian~\cite{tang2023dreamgaussian} utilized a similar approach but trained a surface represented with Gaussian Splatting~\cite{kerbl20233d}
and used SDS instead of the synthesized images directly.
To the best of our knowledge, there are no works on generative \emph{multi-view} 3D reconstruction that address the same problem as ours,
so we compare with the three works described above.
A different work, SparseFusion~\cite{zhou2023sparsefusion}, utilizes a diffusion model for few-view 3D reconstruction, but it is only applicable to class-specific data.

\noindent \textbf{Shape completion} is a vast area of research.
We only discuss recent shape completion methods that leverage generative modeling techniques,
and refer the reader to a survey~\cite{fei2022comprehensive} for a broader review of the topic.
One approach is to use 3D generative models conditioned on the partial shape,
either using GANs~\cite{wu2020multimodal,zhang2021unsupervised,cai2022learning},
transformer~\cite{yu2021pointr,xiang2021snowflakenet,yan2022shapeformer},
or diffusion~\cite{zhou20213d,chu2023diffcomplete,cheng2023sdfusion}.
Typically, these methods are trained on smaller datasets like ShapeNet~\cite{chang2015shapenet} and on synthetically generated inputs.
These two limitations preclude a comparison in our setting, as these methods cannot handle our inputs.
A concurrent work~\cite{kasten2023point} shows nice initial results using SDS.
In an informal comparison, we observe that our method yields higher visual quality, but a systematic comparison is only feasible once the code becomes available.

\section{Method}
\label{sec:method}
We start with the description of the core ideas of NeuS~\cite{wang2021neus} and Score Distillation Sampling~\cite{poole2022dreamfusion} that our method builds on
and then describe the novel components of our method.

\subsection{NeuS}
We choose NeuS~\cite{wang2021neus} as the starting point and the baseline for our work.
NeuS is a volumetric differentiable rendering method that parametrizes the surface using two deep neural networks $f: \mathbb{R}^3 \rightarrow \mathbb{R}$ and $c: \mathbb{R}^3 \times \mathbb{S}^2 \rightarrow \mathbb{R}^3$ that represent the geometry as signed distance function (SDF) and the appearance as radiance field respectively. The surface is defined as the zero-level set of the SDF network.
The color of a pixel rendered along the ray with direction $\bm{v}$ starting at the point $\bm{o}$ is given by 
\begin{equation}
\label{eq:volumetric_rendering}
C(\bm{o}, \bm{v}) = \int_{0}^{+\infty} W(l|f)c(\bm{p}(l), \bm{v})\mathrm{d}l,
\end{equation}
where $\bm{p}(l)=\bm{o}+l\bm{v},\, l > 0$ is a point on the ray,
and $W(l|f)$ is a weight function that depends on the SDF network.

To fit the model to a set of images, one optimizes the photometric term \(\mathcal{L}_\mathrm{c} = \sum_{k} \|C(\bm{o}_k, \bm{v}_k) - C_{k}\|_{1} / K\)
over the set of observed pixel values $\{C_{k}\}$ and the respective camera positions \(\bm{o}_k\) and view directions \(\bm{v}_k\).
The loss function of NeuS additionally includes the mask guidance term \(\mathcal{L}_\mathrm{m}\),
and the Eikonal penalty \(\mathcal{L}_\mathrm{eik}\), intended to assure that the SDF network $f$ represents a valid signed distance function.
We refer the reader to~\cite{wang2021neus} for the definition of these terms.

\subsection{Score Distillation Sampling}
DreamFusion~\cite{poole2022dreamfusion} introduced the concept of score distillation sampling (SDS)
for training neural radiance fields (NeRF) using a pre-trained 2D diffusion model, without any real-world input images.
The idea behind SDS is to pass the image rendered from the NeRF during training with added Gaussian noise
through the denoising diffusion model conditioned on a textual description of the scene,
and to use the output of the model to guide the NeRF towards a representation that better corresponds to the description.

For a 3D scene parameterized by $\theta$, the rendered image $\bm{x}(\theta)$,
and the added Gaussian noise $\boldsymbol{\epsilon}$,
the denoising model $\epsilon_{\phi}$ produces the output
\(\hat{\bm{\epsilon}} = \epsilon_\phi(\bm{y}, t, \alpha_t \bm{x} + \sigma_t \boldsymbol{\epsilon})\),
that is then used to guide the NeRF with the gradient of the virtual SDS loss term
\begin{equation}
    \label{eq:sds_loss}
    \nabla_{\theta} \mathcal{L}_\mathrm{SDS}=\mathbb{E}_{t, \bm{\epsilon}} \left[w(t)(\hat{\bm{\epsilon}} - \bm{\epsilon})\frac{\partial {\bm x}}{\partial \theta}\right],
\end{equation}
where $\bm{y}$ is the embedding of the text prompt,
$t \sim \mathcal{U}(0, 1)$ is the diffusion timestep,
and \(\alpha_t\), \(\sigma_t\), and \(w(t)\) are the weighting factors that depend on the timestep.
Please refer to~\cite{poole2022dreamfusion} for their definition.

\noindent \textbf{SDS for surface completion.}
We employ SDS to guide the formation of the unobserved part of the NeuS surface fitted to a set of images, the viewpoints for which we denote as \emph{visible}.
For this, we define a set of viewpoints directed to the unobserved side of the surface and apply the virtual SDS loss term for these viewpoints,
obtaining the rendered image~\(\bm{x}\) through~\cref{eq:volumetric_rendering}.
The loss function for training NeuS with SDS guidance is given by
\begin{equation}
    \label{eq:neussd_total_loss}
    \mathcal{L}_\mathrm{NeuS+SDS}=\mathcal{L}_\mathrm{c} + \beta \mathcal{L}_\mathrm{m} + \lambda \mathcal{L}_\mathrm{eik} + \gamma \mathcal{L}_\mathrm{SDS}.
\end{equation}
For the SDS implementation, we follow~\cite{poole2022dreamfusion} but use a latent diffusion model, Stable Diffusion 2.1,
so in our formulation of the SDS loss term the rendered image \(\bm{x}\) in~\cref{eq:sds_loss} is replaced with the respective latent code.

\subsection{SDS Surface Shape Guidance}
\label{sec:normal_sds}
Typically SDS is used in conjunction with color renderings. In addition, it can be used to guide other types of renderings, \eg, shaded albedo and textureless renderings to improve geometric details and avoid degenerate flat solutions.
Based on our experiments, we argue that shading is a mere projection of the surface normals leading to information loss.
Instead, we suggest using normal maps directly, which are easy to obtain from neural surfaces by rendering the SDF derivative similarly to~\cref{eq:volumetric_rendering}, via
\begin{equation}
\label{eq:normal_map_rendering}
N(\bm{o}, \bm{v}) = \int_{0}^{+\infty} W(l|f) \nabla f(\bm{p}(l))\mathrm{d}l.
\end{equation}
The normal map is then used instead of the rendered color image~\(\bm{x}\) in~\cref{eq:sds_loss} (or rather its respective latent code) to compute the normal-based SDS loss term \(\mathcal{L}_\mathrm{SDS,N}\).

A normal map with its components mapped to RGB, as in~\cref{fig:fig_teaser}, could be viewed as a textureless render under special lighting conditions (not taking self-occlusion into account) with three distant colored light sources: red, green, and blue.
Thus, normal maps should lie inside the generative manifold of the diffusion model with enough generalization capacity and be a valid input for the SDS algorithm.

The color renderings, despite their redundancy, may still contain important information that is not present in the normal maps,
so we propose to use both the color- and normal-based SDS for training:
\begin{align}
    \label{eq:neussd_total_loss_normals}
  \begin{split}  \mathcal{L}_\mathrm{+Normals}=\mathcal{L}_\mathrm{c} &+ \beta \mathcal{L}_\mathrm{m} + \lambda \mathcal{L}_\mathrm{eik} + \\
    &+ \gamma \mathcal{L}_\mathrm{SDS} + \gamma_{N} \mathcal{L}_\mathrm{SDS,N}.
    \end{split}
\end{align}

Since the diffusion model may have some form of bias and associate particular prompts with particular colors, \eg, \textquote{photo camera} with black color,
we randomly rotate the normals during training to ensure uniform color distribution of the color-mapped normals and avoid implicit bias.

\subsection{Frozen SDS}
\label{sec:frozen_sds}
During the initial phases of learning the 3D representation with SDS the surface is not yet formed and the gradient from the denoising network is highly inconsistent, which leads to artifacts and slow convergence. One way to address this issue~\cite{zhu2023hifa} is to employ multi-step bootstrapping for SDS to make it more consistent.
Instead of computationally expensive multi-step bootstrapping, we propose to fix the additive Gaussian noise $\bm{\epsilon}$ in~\cref{eq:sds_loss} \textquote{frozen}
\begin{equation}
    \label{eq:sds_frozen_loss}
    \nabla_{\theta} \mathcal{L}_\mathrm{SDS,frozen}=\mathbb{E}_{t} \left[w(t)(\hat{\bm{\epsilon}} - \bm{\epsilon}_\mathrm{fixed})\frac{\partial \bm{x}}{\partial \theta}\right].
\end{equation}
Fixing the noise makes the gradient dependent only on the timestep, which drastically reduces the variation of the gradient, and not only leads to faster convergence but also improves the generation results.

In our method, we combine the idea of~\cref{eq:sds_frozen_loss} with the loss function from~\cref{eq:neussd_total_loss_normals}.

\subsection{Multi-view SDS}
\label{sec:mv_sds}
SDS is known to produce inconsistent updates from different viewpoints, leading to the well-known \emph{Janus problem}~\cite{armandpour2023re}.
While this particular issue is not central to our work, the observed and the unobserved parts of the surface still need to be reconstructed consistently.
Previous works have addressed this issue in different ways. For example, RealFusion~\cite{melas2023realfusion} fine-tunes the Stable Diffusion (SD) model using textual inversion~\cite{gal2022image}, which is time-consuming and prone to overfitting.
SparseFusion~\cite{zhou2023sparsefusion} trains a view-conditioned diffusion on a category-specific multi-view dataset. Unfortunately, such data is not widely available and the multi-view model lacks generalization capacity compared to the regular 2D diffusion model.

Instead, we propose to condition the diffusion model on multiple views without any tuning, architecture changes, or training of a different model from scratch. 
We observe that the publicly available SD model, given the prompt of the form \emph{\textquote{renders of the same \textlangle X\textrangle{} from different viewpoints}},
produces a grid of samples that share distinctive stylistic and geometric similarities, although not perfectly aligned (see examples in the supplementary material).
Given this, we propose to apply the SDS loss term to a grid of images compiled of one rendering from the unobserved viewpoint and several renderings from the visible viewpoints, as shown in~\cref{fig:fig_teaser}.
This provides the unobserved part of the surface with a guidance signal consistent with the observed parts.
We note that we use the parts of the grid rendered from the visible viewpoints only for consistency guidance,
and do not propagate the gradient from the SDS loss term to the surface representation through these parts.

We train our complete method using the loss function from~\cref{eq:neussd_total_loss_normals} combined with the ideas of frozen and multi-view SDS.

\begin{table*}[ht]
    \centering
    \begin{tikzpicture}
        \node[anchor=south west,inner sep=0] (image) at (0,0) {\includegraphics[width=\textwidth]{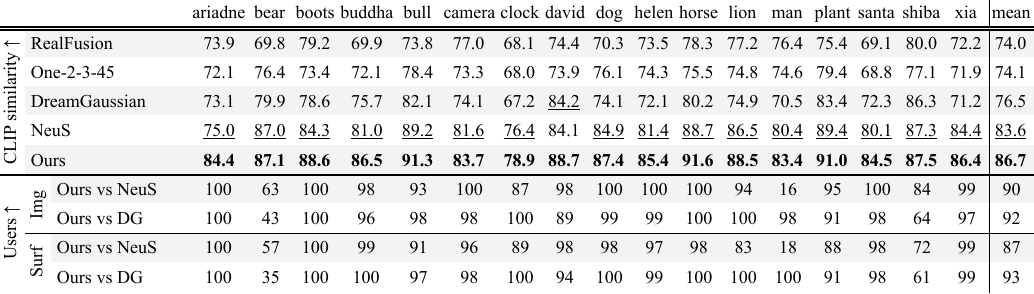}};
        \begin{scope}[shift=(image.north west), x=1pt, y=-1pt, anchor=south west, inner sep=0, font=\small]
            \node at (50, 25.5) {\phantom{n}~\cite{melas2023realfusion}};
            \node at (51, 39.8) {\phantom{5}~\cite{liu2023one}};
            \node at (65.8, 53.5) {\phantom{n}~\cite{tang2023dreamgaussian}};
            \node at (29.2, 67.5) {\phantom{S}~\cite{wang2021neus}};
            \node at (70.5, 95.75) {\phantom{S}~\cite{wang2021neus}};
            \node at (62.5, 109.8) {\phantom{G}~\cite{tang2023dreamgaussian}};
            \node at (70.5, 124) {\phantom{S}~\cite{wang2021neus}};
            \node at (62.5, 138) {\phantom{G}~\cite{tang2023dreamgaussian}};
        \end{scope}
    \end{tikzpicture}
    \caption{
    \textbf{Quantitative comparison.}
    Top: CLIP similarity, the \textbf{best} and \underline{second best} results are highlighted.
    Bottom: results of the user study for the two questions described in the main text. The values represent the percentage of subjects that prefer our method over the respective competitor.
    }
    \label{tab:tab_metrics}
\end{table*}

\section{Experiments}
\label{sec:experiments}
\subsection{Evaluation setup}
\noindent \textbf{Data.}
We evaluate our method using the BlendedMVS~\cite{yao2020blendedmvs} semi-synthetic dataset for multi-view stereo reconstruction,
consisting of 117 scenes including small objects, statues, and architecture.
Each scene usually contains a single salient object, the data for which consists of the reference 3D surface and the images captured from 30-200 different directions. 
We chose BlendedMVS because it provides viewpoints all around the object, required for evaluation in our setting, in contrast to other widely-used datasets for multi-view reconstruction like DTU~\cite{jensen2014large,aanaes2016large}.

\noindent \textbf{Data preprocessing.}
We picked 17 scenes from the dataset for our experiments.
For each scene, we manually chose the observed side of the object, the viewpoints from the dataset that mostly capture only the observed side,
and the guidance viewpoints that mostly capture only the unobserved side, 5-15 viewpoints in each set.
To obtain the reference surface for evaluation, we only kept the salient object and manually removed the unrelated environment.
We also used these surfaces to obtain image object masks required for the methods.

\noindent \textbf{Implementation details.}
We took the implementation of the volumetric rendering from NeuS~\cite{wang2021neus}
and an open-source diffusion model Stable Diffusion 2.1 from HuggingFace~\cite{rombach2022high, stablediffusion_url}.
We manually picked the text prompt for the diffusion model per scene and we refer to these prompts in the supplementary.

At each iteration, we randomly pick a viewpoint from the combined set of visible and unobserved views
and compute the regular NeuS loss for the visible viewpoints, and the SDS-guided loss for the unobserved ones.
To compute the SDS loss, we render dense images/normal maps at \(64\times 64\) resolution, to reduce the computational cost, and resize them to \(512\times 512\) resolution before feeding them to the diffusion model.
We alternate between the single- and multi-view SDS to obtain more comprehensive guidance.
For the multi-view SDS, we store the normal maps for the observed part of the object obtained at the last three iterations with visible viewpoints
and combine them with the normal map from the current unobserved viewpoint into a \(2\times 2\) grid.
For every 4-th iteration of the single- or multi-view SDS, we perform the regular color-based SDS.

We kept the implementation of NeuS and its hyperparameters unchanged.
For the SDS loss, we picked the weight around $10^{-5}$, with the optimal exact value depending on the size of the unobserved part of the object.
We used the classifier-free guidance scale of 100 and a timestep uniformly sampled from \([0, 0.5]\) with 1 being the possible maximum.
For each scene, we trained the model for 300k iterations, which took 28 hours on a single A100 NVIDIA GPU,
13 of which were spent on Stable Diffusion inference.

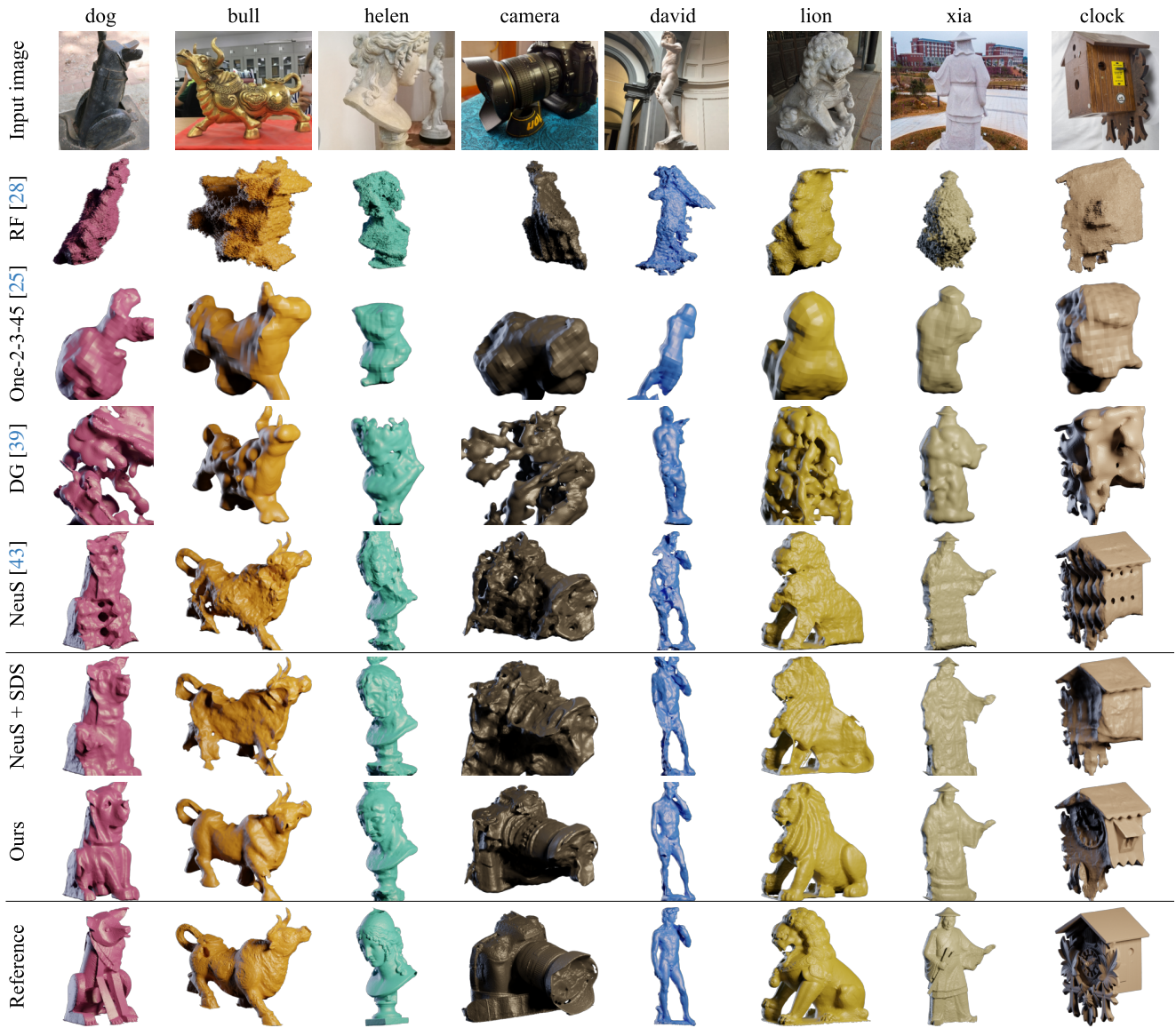
\begin{figure*}[h]
    \centering
    \begin{tikzpicture}
        \node[anchor=south west,inner sep=0] (image) at (0,0) {\includegraphics[width=\textwidth]{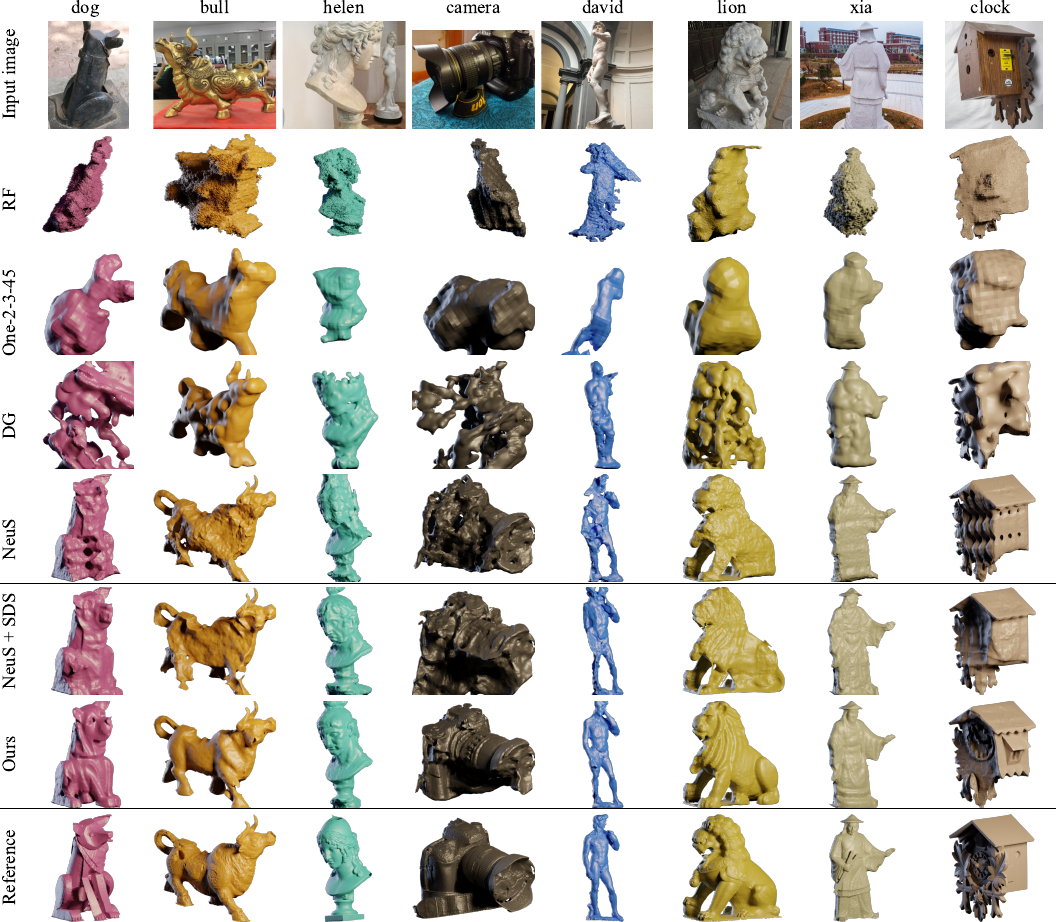}};
        \begin{scope}[shift=(image.north west), x=1pt, y=-1pt, anchor=south west, inner sep=0, font=\small]
            \node at (1.3, 94) {\rotatebox{90}{\phantom{F}~\cite{melas2023realfusion}}};
            \node at (1.3, 131.5) {\rotatebox{90}{\phantom{5}~\cite{liu2023one}}};
            \node at (1.3, 201) {\rotatebox{90}{\phantom{G}~\cite{tang2023dreamgaussian}}};
            \node at (1.3, 249.5) {\rotatebox{90}{\phantom{S}~\cite{wang2021neus}}};
        \end{scope}
    \end{tikzpicture}
    \caption{
        \textbf{Qualitative comparison.}
        The first row shows the image from the visible set used by the competing single-view methods as input. The next four rows show the results of the three generative single-view competitors: RealFusion (RF), One-2-3-45, and DreamGaussian (DG), and the reconstructive competitor (NeuS). The next two rows show the results of the multi-view baseline (NeuS + SDS) and of our full method. The last row depicts the reference surface. All meshes are rendered from the unobserved side.
    }
    \label{fig:fig_qualitative}
\end{figure*}

\noindent \textbf{Competitors.}
As a representative of purely reconstruction-based methods, we selected NeuS~\cite{wang2021neus} because it constitutes the starting point for our work. Even though there are many follow-up works to NeuS, all of them are expected to behave similarly in reconstructing the unseen parts of the object.
We used the official implementation and obtained the results for NeuS using the images and masks for the whole set of visible viewpoints.

As representatives of generative methods we selected RealFusion~\cite{melas2023realfusion}, One-2-3-45~\cite{liu2023one}, and DreamGaussian~\cite{tang2023dreamgaussian},
that condition a diffusion model on a single input image and utilize this model to guide the surface reconstruction process.
We used the official implementation of all methods and obtained the results for the input image
that we manually selected from the visible viewpoints so that it provides the most context about the unobserved part of the object.
We additionally provided the methods with the ground truth image object mask and made them aware of the ground truth input camera position \wrt~the object,
as we describe in the supplementary material.
We initialized the prompt for RealFusion with the same prompt we used for our method.

\noindent \textbf{Metrics.}
We evaluate the quality of the results using CLIP similarity and a perceptual user study.
For CLIP similarity, we render the reconstructed and the reference surfaces using viewpoints distributed uniformly around the object, compute the cosine similarity between the respective CLIP~\cite{radford2021learning} image embeddings, and report the average value across all views.

For the user study, we conducted an online survey in which each participant was asked to evaluate the results of our method, NeuS~\cite{wang2021neus}, and DreamGaussian~\cite{tang2023dreamgaussian} from the unobserved side using two tasks:
picking the reconstruction that corresponds better to an input image, and picking the reconstruction that is more similar to the shown reference surface.
We report the responses of 130 participants.

We also compared the accuracy of the reconstructed surface. Our method accurately reconstructs the observed part of the object, as we show in~\cref{fig:fig_teaser}.
The accuracy computed for the whole surface is, as expected, similar for all methods.
We report more on this in the supplementary material.

\subsection{Quantitative and Qualitative Results}
We show the quantitative comparison of the methods in~\cref{tab:tab_metrics} and the qualitative comparison in~\cref{fig:fig_qualitative}.
We first discuss the quantitative results using CLIP and then the qualitative results jointly with the user study.

Our method consistently outperforms both the generative competitors and the baseline NeuS \wrt~CLIP similarity. This demonstrates that we can reconstruct the observed parts of the surface well, but at the same time achieve more plausible generative shape completions for the unobserved parts.
We note that we measure CLIP similarity for both the observed and unobserved parts, so the shown numbers are a blend between reconstruction and generation quality. If we use CLIP similarity only for the generated parts, our method has an even bigger advantage compared to NeuS, as we show in the supplementary material.

The user study, evaluating the generation of the unobserved parts only, shows a very strong preference for our method on most scenes.
We obtained close to 100 percent preference votes for the majority of the scenes. While a preference score of 100 percent is unusual, one can confirm in~\cref{fig:fig_qualitative,fig:fig_teaser} that the reconstruction quality of the competitors is low for these scenes. There are multiple reasons why competitors struggle to generate plausible results for the unobserved parts. NeuS is not a generative method and generally produces overly smooth shape completions that cannot be judged to be realistic.
The generative single-view competitors seem to be strongly dependent on an aligned frontal view. If the single input view does not contain the most salient object parts, the output degenerates.
We additionally discuss the challenges of our setting for these methods in the supplementary material.

\begin{figure}[t]
    \centerline{\includegraphics[width=\columnwidth]{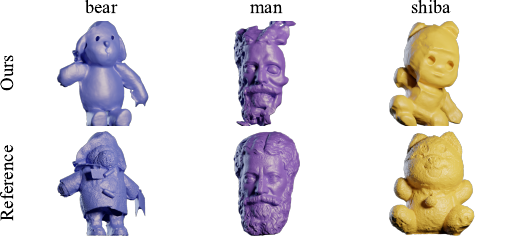}}
    \caption{Scenes with lower user scores for our method.}
    \label{fig:fig_fails}
\end{figure}

The scenes that show lower user scores for our method are \emph{bear}, \emph{man}, and \emph{shiba}. The qualitative results shown in~\cref{fig:fig_fails}
confirm that our reconstructions for these scenes are indeed worse. Our conjecture for this problem is that our prompts \textquote{soft toy}, \textquote{antique death mask}, and \textquote{asian toy}, respectively, are ill-suited for these scenes because the Stable Diffusion 2.1 interprets them differently than intended.
For the \emph{bear} it tends to generate the forward side of the regular teddy bear and for the \emph{shiba} it associates the chosen prompt with some plastic toy. We generally opted for using simple and short prompts and chose not to spend a lot of time fine-tuning them for our method. While it could be possible to get better results when trying many different prompts, we also observed in some experiments that seemingly very fitting and elaborate prompts can lead to poor results.

For the \emph{man}, which is the face of a statue for which we picked one half of the face as the observed part, we notice a conflict between reconstruction and generation with the prompt. While both sides of the face should be similar to the visible part, the reconstruction and the generation cannot agree, and the generated information overrides the information from the reconstructed part, rather than the other way around.
We find that quite intriguing because for other examples, \eg, \emph{bull}, \emph{helen}, or \emph{santa}, we noticed that symmetry is actually helpful for generation,
and that the SDS loss helps to propagate information from the visible part to the unobserved part without creating a conflict.

Overall, we observe that single-view methods are limited by the fact that they cannot consider all views from the visible set as input.
As there is no existing method available to tackle exactly the same problem as our method, we chose to implement a combination of NeuS with color-based SDS as an additional multi-view baseline. The advantage of this method is that it also serves as an ablation for our work as it is directly comparable. We compare with NeuS + SDS in the ablation study described in the next subsection.
We also evaluated the code of multiple other single-view methods, attempting to extend them to the multi-view case. However, this requires a major engineering effort that exceeds a simple change. It would require an adaption of the prompting strategy and prompt generation, viewpoint generation, and searching for new hyperparameters.

\begin{figure*}[h]
    \centering
    \begin{tikzpicture}
        \node[anchor=south west,inner sep=0] (image) at (0,0) {\includegraphics[width=\textwidth]{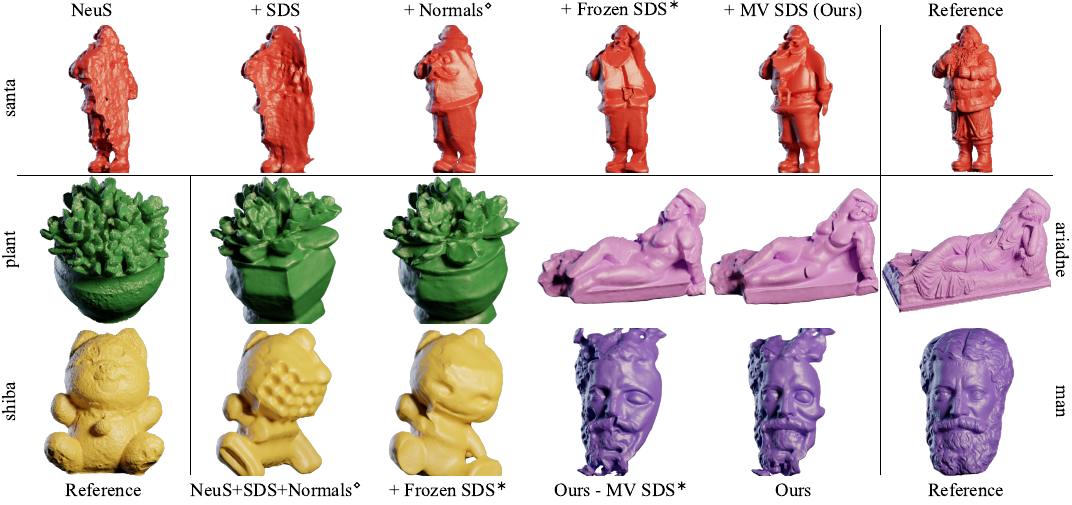}};
        \begin{scope}[shift=(image.north west), x=1pt, y=-1pt, anchor=south west, inner sep=0, font=\small]
            \node at (47.5, 9.5) {\phantom{S}~\cite{wang2021neus}};
        \end{scope}
    \end{tikzpicture}
    \caption{
        \textbf{Ablation study.} The first row shows how our method builds on NeuS by successively adding color image-based SDS guidance, normal maps, frozen SDS, and multi-view SDS guidance. The second and third rows show partial ablation studies for four scenes. We illustrate the effect of freezing the noise on the left, and the effect of multi-view SDS guidance on the right. The superscript markers \(\diamond\) and \(\ast\) mark the same versions of the method.
    }
    \label{fig:fig_ablation}
\end{figure*}

\subsection{Ablation study}
For ablation, we start from the baseline NeuS method and add SDS as generative guidance to establish a multi-view baseline. Then, we gradually add components from~\cref{sec:method}, namely normal maps, frozen SDS, and the multi-view SDS.
We provide some qualitative visual analysis of the ablation study at the bottom of~\cref{fig:fig_qualitative} and in~\cref{fig:fig_ablation},
and show the full comparison in the supplementary material.

We observe that the most significant component of our method is the introduction of normal maps in conjunction with SDS, which greatly improves the quality of the produced surfaces for all scenes compared to using color renderings only. The normal map supervision seems to work better with viewpoints sampled from only a side of the object compared to SDS on color images, and is therefore more suitable to mix reconstruction and generative losses.

The addition of the frozen SDS mainly speeds up the convergence of the method and in some cases slightly improves the results for large unobserved parts, as we show for the \emph{santa}, \emph{plant}, and \emph{shiba}, making them more consistent with the visible part of the surface.
We hypothesize that the static noise map decreases the variability in the learning signal and thereby consistently guides the formation of the unseen part of the surface.

Finally, the multi-view SDS makes the generated part of the surface more consistent with the observed part over the number of scenes. 
Its effectiveness seems to largely depend on the stylistic similarity of the visible images with the unobserved images.
For example, it makes the body pose of the \emph{ariadne} more consistent with the visible images and more similar to the ground truth pose, makes the \emph{man} more symmetric, and for the \emph{santa} greatly improves the overall quality of the surface and makes the figure aligned with the overall pose.

We also tried to evaluate the differences between the different versions of the method in the ablation study using CLIP similarity. The differences are generally too subtle to be picked up by CLIP features and therefore the methods all show similar results within a narrow range. We show these results in the supplementary material.

\section{Conclusions}
\label{sec:conclusions}

We proposed a novel framework, called NeuSD. It mixes traditional NeuS-based 3D shape reconstruction with generative completion. NeuSD is especially useful for cases where part of a shape is observed by multiple images, and another part of the shape is not observed. It creates a plausible completion of the shape that is semantically meaningful and consistent with the observed parts. The main components of our approach are: 1) surface diffusion guidance, 2) freezing noise, and 3) multi-view SDS. Our results outperform the current state of the art in qualitative and quantitative metrics.
We also would like to discuss two limitations of our work. First, the computation times are still fairly high, compared to some of the fastest known methods that just use a single forward pass through a network to generate a shape.
Second, we do not use a separately trained diffusion network to generate multiple-view images like~\cite{liu2023one}. This possibly limits the quality of the results. However, we do not think that current multi-view datasets are good enough for real-world reconstruction.
In future work, we would like mainly to address the inference time problem to investigate faster approaches for mixing surface reconstruction and generative modeling. In addition, we would like to extend our framework to complete scenes containing multiple objects.
\vspace{7pt plus 1pt minus 1pt}
\paragraph{Acknowledgements.}
The authors acknowledge the use of Skoltech supercomputer Zhores~\cite{zacharov2019zhores} for obtaining the results presented in this paper.
The authors from Skoltech were supported by the Analytical center under the RF Government
(subsidy agreement 000000D730321P5Q0002, Grant No. 70-2021-00145 02.11.2021).

\vfill

    \clearpage
\maketitlesupplementary

In~\cref{sec:sup_data} we describe our test data in more detail.
In~\cref{sec:sup_implementation} we provide additional details of
selection of textual prompts for the text-to-image diffusion model used in our method,
selection of the timestep for the diffusion model,
and testing details for the competing methods.
In~\cref{sec:sup_evaluation_details} we describe our quantitative evaluation in detail.
In~\cref{sec:sup_comparison,sec:sup_challenges} we provide
the complete comparison of our method with the competing methods,
and discuss the challenges of our setting for the competing methods.
In~\cref{sec:sup_ablation} we show the complete set of qualitative and quantitative results of our ablation study.
In~\cref{sec:sup_multi_view,sec:sup_frozen_sds} we additionally discuss the premise of multi-view SDS and theoretical aspects of frozen SDS.

\section{Data selection}
\label{sec:sup_data}
We evaluated our method on 17 scenes from the BlendedMVS~\cite{yao2020blendedmvs} dataset.
For each scene, we used two sets of viewpoints:
the \emph{visible} viewpoints with the respective input images that capture the observed part of the surface,
and the guidance viewpoints that our method uses to apply Score Distillation Sampling (SDS)~\cite{poole2022dreamfusion}.
We show the images for the visible viewpoints in~\crefrange{fig:sup_fig_visible_1}{fig:sup_fig_visible_3}.
We selected both sets of viewpoints from all viewpoints in the dataset manually using the following algorithm.

First, we chose the unobserved side for each scene.
In general, we chose the side of the surface that is intuitively harder to recover given only the information about the opposing side,
such as the face of a statue or a figurine.
To increase the diversity of the test data and to evaluate the ability of our method to exploit bilateral symmetry,
we also chose a lateral unobserved side for some scenes,
specifically, for \emph{bull}, \emph{camera}, \emph{helen}, \emph{horse}, \emph{lion}, and \emph{man}.
Next, we picked 5--15 visible viewpoints
that on the one hand capture the chosen unobserved side as little as possible,
and on the other hand provide a sufficient parallax for multi-view stereo reconstruction of the visible side of the surface.
Finally, we picked 5--15 guidance viewpoints based on similar considerations for swapped sides of the surface,
\ie, directed to the unobserved side.

We emphasize that we picked the data for our experiments solely based
on our intuition about the difficulty of completion of the unobserved side,
and not based on the performance of our method.
We picked the viewpoints once and kept them fixed in all experiments.
We note that the described algorithm could be formally implemented in software,
similarly to view selection used for multi-view surface reconstruction (\eg,~\cite{schoenberger2016pixelwise} or~\cite{yao2018mvsnet}, Section~4.1).

\begin{table}[t]
    \centerline{\includegraphics[width=\columnwidth]{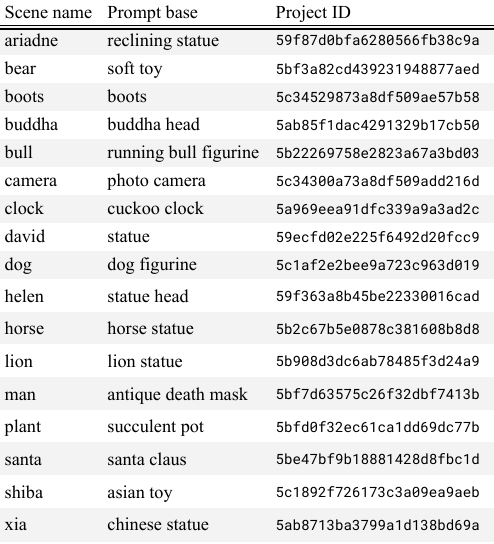}}
    \caption{\textbf{Textual prompts} that we used for the scenes in our experiments and their respective project IDs in the BlendedMVS dataset.}
    \label{tab:sup_tab_prompts}
\end{table}

\section{Implementation details}
\label{sec:sup_implementation}

\noindent \textbf{Prompts.}
To apply SDS guidance in our method we used Stable Diffusion 2.1 conditioned on a textual prompt.
We manually picked simple and short prompts based on the input images for each scene:
we show these prompts in~\cref{tab:sup_tab_prompts}.
We used them as is for the NeuS+SDS baseline and as the initialization for textual inversion in RealFusion~\cite{melas2023realfusion},
and for our complete method with multi-view SDS we substituted these prompt bases with
\textquote{renders of the same ( \emph{prompt base} ) from different viewpoints}.

\noindent \textbf{Timestep interval for SDS.}
The effect of SDS depends on the timestep parameter \(t\) of the underlying denoising diffusion model.
This parameter controls the magnitude of the added noise and the scale of the guidance updates:
greater values lead to lower-frequency updates and smaller values lead to higher-frequency updates~\cite{zhu2023hifa}.
Originally, DreamFusion~\cite{poole2022dreamfusion} suggested to sample \(t\) from the \([0,1]\) interval.
Further work~\cite{zhu2023hifa} proposed to gradually decrease the upper threshold of the interval to improve the sharpness of the results.
Unlike these two works which aim at generation from scratch, we focus on shape completion and want to keep a large part of the surface intact.
Therefore, we cut the upper threshold, finding that the interval of \([0, 0.5]\) is sufficient for our needs.

\noindent \textbf{Testing competitors.}
For our comparisons we selected three generative methods,
namely RealFusion~\cite{melas2023realfusion}, One-2-3-45~\cite{liu2023one}, and DreamGaussian~\cite{tang2023dreamgaussian}.
All three methods take as input a single image,
and in their official implementations assume some default camera model for this image
and only require the elevation angle \wrt~the reconstructed object to constrain the camera position.
RealFusion additionally assumes the input image to be captured from the frontal side of the object,
and uses this assumption to make view-dependent textual prompts for the text-to-image model.

For a fair comparison of these methods with NeuS~\cite{wang2021neus} and our method
that both use ground-truth camera models and positions,
we modified the implementations of the generative methods to account for these camera parameters.
First, we replaced the default camera models with the ground-truth ones.
Next, we calculated the elevation angles for each scene from the ground-truth camera positions,
setting the vertical axis of the object using the floor part of the reference mesh.
Finally, for RealFusion we additionally calculated the horizontal position of the input image \wrt~the object,
setting the forward direction of the object manually.

We found that RealFusion tends to produce disconnected parts of the scene.
Therefore, we additionally postprocessed the surfaces produced by all methods keeping only the largest connected component,
which is also a common post-processing step in the literature on neural surface reconstruction.

\section{Evaluation details}
\label{sec:sup_evaluation_details}

\noindent \textbf{CLIP similarity.}
We used CLIP similarity to evaluate the perceptual quality of the whole surface,
and additionally, to evaluate the unobserved part only.

To evaluate the whole surface, we rendered the reconstructed and the reference surfaces using viewpoints distributed uniformly around the object,
computed the cosine similarity between the respective CLIP~\cite{radford2021learning} image embeddings, and took the average value across all views.
We used {ViT-L/14} CLIP model~\cite{clip_hf}.

To evaluate the unobserved part only,
we discarded the viewpoints for which more than \(\nicefrac{1}{3}\) of the rendering of the reference surface corresponded to the visible part.
To define the visible part of the reference surface formally, we tested the visibility of each triangle of the reference mesh
from the visible viewpoints (shown in~\crefrange{fig:sup_fig_visible_1}{fig:sup_fig_visible_3}) taking self-occlusions into account.
We marked a triangle visible if its center was traceable without occlusions from at least 3 viewpoints.

\begin{figure}[ht]
    \centerline{\includegraphics[width=\columnwidth]{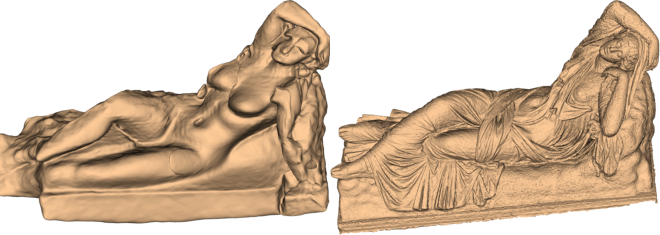}}
    \caption{\textbf{Example of renderings used for CLIP similarity.}}
    \label{fig:sup_fig_clip_renders}
\end{figure}

In~\cref{fig:sup_fig_clip_renders} we show an example of a pair of renderings that we compared using CLIP.
To obtain such renderings, we used an orthographic camera
and a simple diffuse shader without shadow mapping and with a single directional light source positioned behind the camera.
A particular challenge of the CLIP similarity is its sensitivity and ability to distinguish differences in the geometry. Empirically, we observed that the sensitivity of CLIP similarity is higher if the surface is rendered in some color instead of grayscale. We also observed that there is no strong dependence on the actual color as long it is not gray, so we used the color shown in~\cref{fig:sup_fig_clip_renders} in all computations with CLIP.

\noindent \textbf{Geometric surface quality.}
To evaluate the geometric surface quality, that we report further in~\cref{sec:sup_comparison},
we used the F-score computed for the whole surface,
and additionally, the recall computed for the visible part only.
We calculated these metrics similarly to how they are calculated in benchmarks on multi-view
surface reconstruction~\cite{Knapitsch2017,schoeps2017cvpr,voynov2023multi}, as we describe below.

First, to bring all scenes to the same scale,
we transformed the reconstructed and the reference surfaces so that the reference surface is fitted into a unit sphere.
Then, to prevent an uneven contribution of different parts of the surface to the metric,
we resampled both surfaces uniformly with a sufficiently high resolution, specifically 0.003 (without changes to the geometry).
After that, we calculated the precision of the reconstructed surface as the fraction of samples on this surface with the distance to the reference below a conservative threshold of 0.02
(which is an order of magnitude higher than what is usually used in benchmarks);
we calculated the recall as the fraction of samples on the reference surface
with the distance to the reconstructed surface below the same threshold.
Finally, we computed the F-score for the whole surface as the harmonic mean of the precision and recall.

To evaluate only the visible part of the surface,
we marked it as we formally described above,
and calculated the recall only for the samples on the visible part.
Since the calculation of the precision \wrt~an incomplete visible part of the surface is ambiguous we only report the recall.

\begin{table*}[t]
    \centering
    \begin{tikzpicture}
        \node[anchor=south west,inner sep=0] (image) at (0,0) {\includegraphics[width=\textwidth]{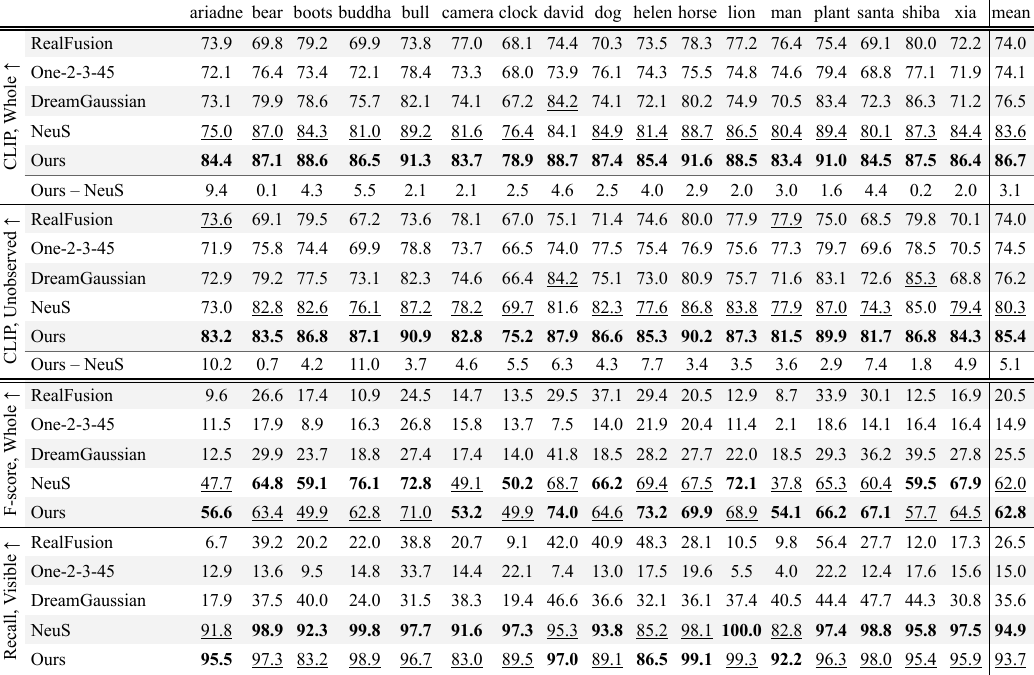}};
        \begin{scope}[shift=(image.north west), x=1pt, y=-1pt, anchor=south west, inner sep=0, font=\small]
            \foreach \yshift in {0,85,169.5,240}{
                \node at (50, 25.5 + \yshift) {\phantom{n}~\cite{melas2023realfusion}};
                \node at (51, 39.8 + \yshift) {\phantom{5}~\cite{liu2023one}};
                \node at (65.8, 53.7 + \yshift) {\phantom{n}~\cite{tang2023dreamgaussian}};
                \node at (29.2, 67.7 + \yshift) {\phantom{S}~\cite{wang2021neus}};
            }
        \end{scope}
    \end{tikzpicture}
    \caption{
    \textbf{Quantitative comparison.}
    Top: CLIP similarity for the whole surface and for the unobserved side only.
    \textquote{Ours -- NeuS} is the difference between the respective values.
    Bottom: the F-score for the whole surface and the recall for the visible side only.
    The \textbf{best} and \underline{second best} results are highlighted.
    }
    \label{tab:sup_tab_metrics}
\end{table*}

\section{More quantitative and qualitative results}
\label{sec:sup_comparison}

We show an additional qualitative comparison of the methods in~\cref{fig:sup_fig_qualitative_other,fig:sup_fig_qualitative_vis_other},
with the surfaces rendered from the unobserved side and from the observed side respectively;
in~\cref{fig:sup_fig_qualitative_vis_main} we show the renderings from the observed side for the scenes shown in Figure~2 in the main text.
In~\cref{tab:sup_tab_metrics} we show an additional quantitative comparison,
using CLIP similarity for the whole surface and the unobserved side only,
the F-score for the whole surface, and the precision for the visible side only.

Our method consistently outperforms all competitors \wrt~CLIP similarity computed for both the whole surface and the unobserved side only,
as we show at the top of~\cref{tab:sup_tab_metrics}.
The advantage of our method compared to NeuS is even bigger for the unobserved side
(the absolute values for both methods are lower since we exclude the more accurately reconstructed visible side from averaging).

In general, we noticed that CLIP similarity of shaded surface renderings is not always consistent with human perception.
While being widely used as a quality measure for colored renderings,
it on the one hand is sometimes not very sensitive to the differences between shaded renderings,
and on the other hand, is sometimes unstable.
For example, the CLIP similarity for the unobserved side of \emph{david} is better for DreamGaussian than for NeuS,
but the result is qualitatively more similar to the reference for NeuS.
We also experimented with other perceptual metrics, namely LPIPS~\cite{zhang2018perceptual} and SSIM~\cite{wang2004image},
but they rather picked up local differences while we were interested in a more global semantic similarity.
Therefore, we additionally evaluated the perceptual quality with a user study, as we describe in the main text.

As for the geometric surface quality, our method accurately reconstructs the visible part of the surface, on par with NeuS,
as we show in~\cref{fig:sup_fig_qualitative_vis_other,fig:sup_fig_qualitative_vis_main},
and confirm with the recall at the bottom of~\cref{tab:sup_tab_metrics}.
The F-score computed for the whole surface shows mixed results
since both methods reconstruct the unobserved part with an arbitrary geometric alignment to the reference.
Our method generates a plausible completion, which does not have to be perfectly aligned,
while NeuS smoothly connects the edges of the visible part.

The generative single-view methods struggle in our setting.
Usually, they are only able to reconstruct the general shape of the object,
as shown in~\cref{fig:sup_fig_qualitative_vis_other,fig:sup_fig_qualitative_vis_main}.
We discuss the challenges of our setting for these methods below.

\begin{figure}[t]
    \centerline{\includegraphics[width=\columnwidth]{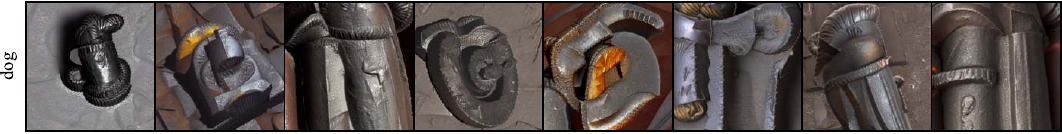}}
    \centerline{\includegraphics[width=\columnwidth]{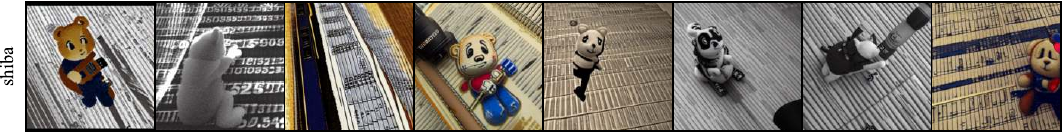}}
    \caption{\textbf{Images generated with Stable Diffusion} 1.5 for textual embeddings estimated in experiments with RealFusion.}
    \label{fig:sup_fig_rf}
\end{figure}
\begin{figure}[t]
    \centerline{\includegraphics[width=\columnwidth]{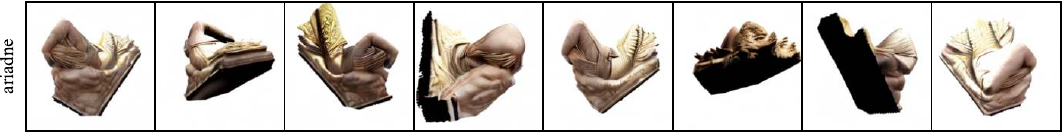}}
    \centerline{\includegraphics[width=\columnwidth]{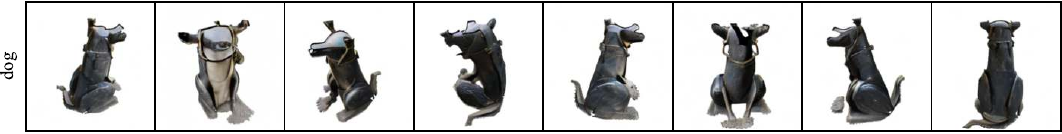}}
    \centerline{\includegraphics[width=\columnwidth]{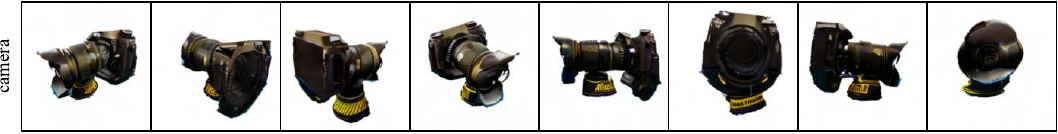}}
    \centerline{\includegraphics[width=\columnwidth]{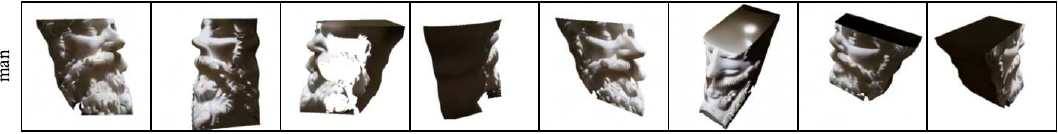}}
    \caption{\textbf{Images generated with Zero123} in experiments with One-2-3-45.}
    \label{fig:sup_fig_one23}
\end{figure}

\section{Discussion of results of competitors}
\label{sec:sup_challenges}

\noindent \textbf{RealFusion}
trains an Instant NGP~\cite{muller2022instant} representation of the scene
that it supervises via SDS using text-to-image Stable Diffusion 1.5 model.
It estimates the textual embedding, to condition the model on, from the input image using textual inversion~\cite{gal2022image}.

In~\cref{fig:sup_fig_rf} we show some examples of images generated with Stable Diffusion 1.5 for the textual embeddings estimated by RealFusion for our test scenes.
We observe that for some scenes the generated images are inconsistent with the input view,
and hypothesize that in our setting, with limited information about the whole surface in the input view,
the textual inversion produces incoherent textual embeddings, which leads to degraded results of RealFusion.
Additionally, the Instant NGP representation of the surface lacks sufficient constraints on its level set,
which leads to further degradation of the results.

\begin{figure}[t]
    \centerline{\includegraphics[width=\columnwidth]{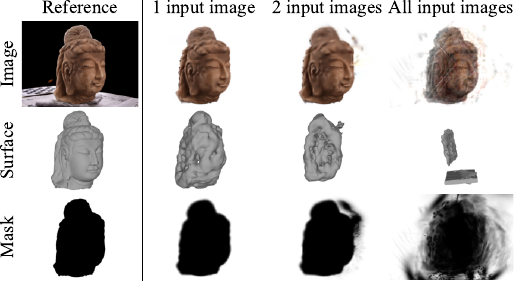}}
    \caption{
        \textbf{Adaptation of DreamGaussian to multi-view setting.}
        We show the color renderings from a fitted model for different numbers of input real-world images in the first row,
        the respective surfaces in the second row,
        and the density renderings from the model in the last row.
        In the first column, we show the reference data.
        For clarity, we show the results for a frontal viewpoint, not used in our main experiments.
    }
    \label{fig:sup_fig_mv_dg}
\end{figure}

\noindent \textbf{One-2-3-45}
trains a SparseNeuS~\cite{long2022sparseneus} representation of the scene
that it supervises with synthetic images generated using Zero-123~\cite{Liu_2023_ICCV} model,
conditioned on the input real-world image and sampled camera poses for the synthetic images.

In~\cref{fig:sup_fig_one23} we show some examples of images produced by Zero-123 for our test scenes.
We observe that for some scenes the generated images lack multi-view consistency,
which inevitably leads to degradation of the 3D surface fitted to them.
The authors of One-2-3-45 note that the quality of the results produced by their method is
\textquote{often limited by the multi-view generation, since there are 3D inconsistencies among the predicted views},
and show a failure case for their method in a setting similar to ours, where the input view contains insufficient information about the whole surface.
Additionally, we observe that for some scenes the generated images, while being rather consistent, represent an incorrect 3D shape.

\noindent \textbf{DreamGaussian,}
similar to One-2-3-45, also uses Zero-123 to estimate multi-view information for the scene from a single input view.
In contrast to One-2-3-45, it does not train the 3D representation of the scene on the generated images directly but instead uses SDS guidance.
This presumably still leads to the same effects related to multi-view inconsistency of the estimation.

As the 3D representation, it uses Gaussian Splatting~\cite{kerbl20233d},
with which we associate the \textquote{hollowness} of the extracted surfaces.
We note that the results that we obtained with this method are qualitatively similar to the results previously obtained by
others~\cite{dg_gh_issue1,dg_gh_issue2,dg_gh_issue3,dg_gh_issue4}.

\noindent \textbf{Multi-view DreamGaussian.}
For the single-view method that arguably produced the best results in our experiments, namely DreamGaussian,
we tried to make an extension to our multi-view setting.
For this, we followed a discussion on GitHub on a similar topic~\cite{multiview_dreamgaussian}.
The modified multi-view version produces a worse surface than the single-view baseline,
as we show in the first two rows of~\cref{fig:sup_fig_mv_dg}.
We confirmed that the model fits the added real-world images one by one,
but when the number of images increases the quality of the result degrades.
We observe that the addition of real-world input images leads to the emergence of gaussian-splats with a nonzero density and the color of the background,
floating around the object, as we show in the last row of~\cref{fig:sup_fig_mv_dg}.
This is one of the possible reasons for the degradation of the surface.
Overall, we hypothesize that the values of the hyperparameters of the method picked for the single-view setting, such as the training parameters of Gaussian Splatting, the parameters of the mesh extraction, or the weights of loss terms, are not suitable for the multi-view setting.

Below, we compare to another multi-view baseline, namely a combination of NeuS with color-based SDS.

\section{More ablation study}
\label{sec:sup_ablation}

We show the qualitative results of the ablation study for all scenes in~\cref{fig:sup_fig_ablation_main,fig:sup_fig_ablation_other}.
In~\cref{tab:sup_tab_ablation} we show the respective quantitative comparison,
using CLIP similarity for the whole surface and the unobserved side only.

Our complete method consistently improves upon the baseline NeuS+SDS, which uses multiple real-world input images.
The addition of the frozen SDS and the multi-view SDS improves the results in several cases qualitatively,
but their effects are not picked up by CLIP similarity on average.

\section{Premise of multi-view SDS}
\label{sec:sup_multi_view}
\begin{figure}[ht]
    \centerline{\includegraphics[width=\columnwidth]{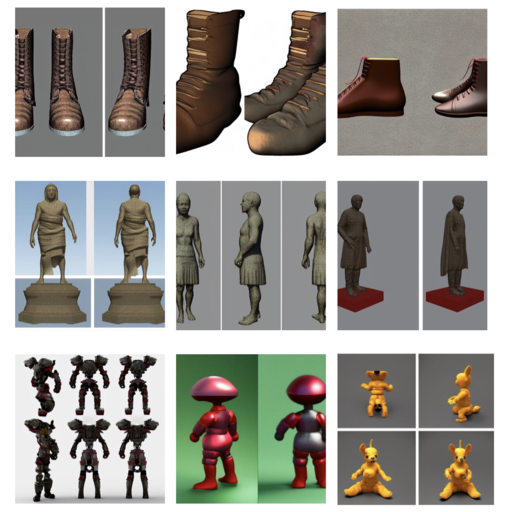}}
    \caption{\textbf{Images generated with Stable Diffusion} for the prompt \textquote{renders of the same boots / statue / toy from different viewpoints}.}
    \label{fig:sup_fig_multi_view_samples}
\end{figure}

\begin{figure*}[p]
    \centerline{\includegraphics[width=\textwidth]{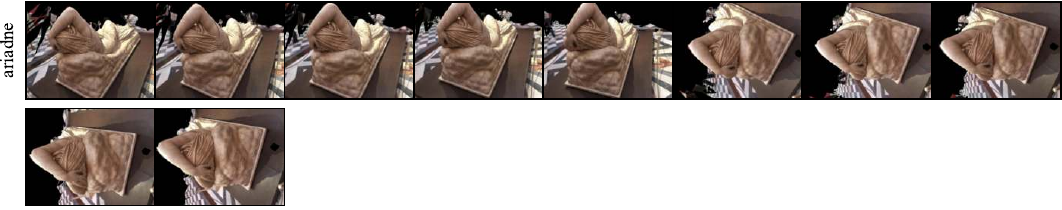}}
    \centerline{\includegraphics[width=\textwidth]{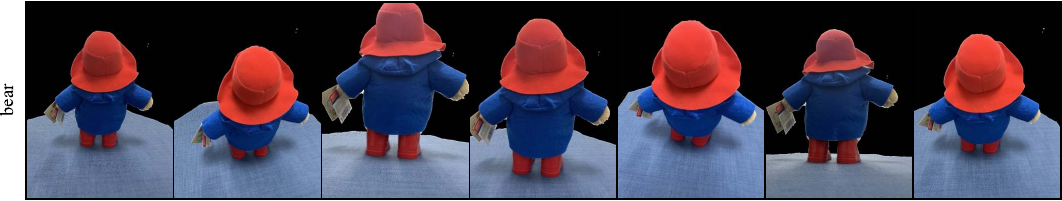}}
    \centerline{\includegraphics[width=\textwidth]{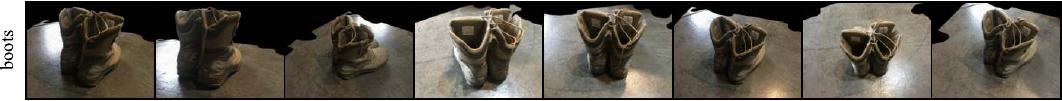}}
    \centerline{\includegraphics[width=\textwidth]{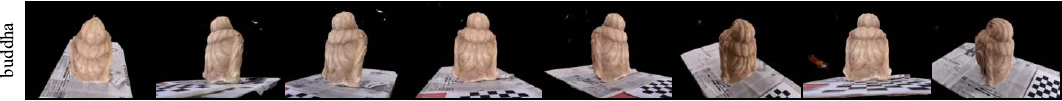}}
    \centerline{\includegraphics[width=\textwidth]{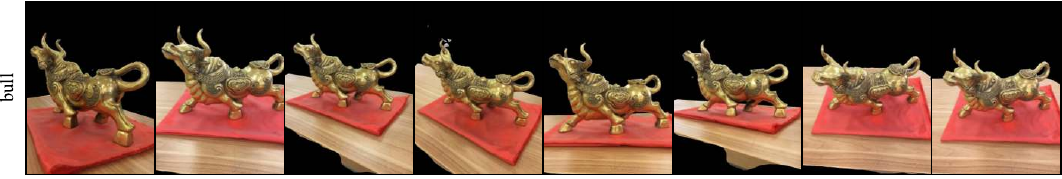}}
    \centerline{\includegraphics[width=\textwidth]{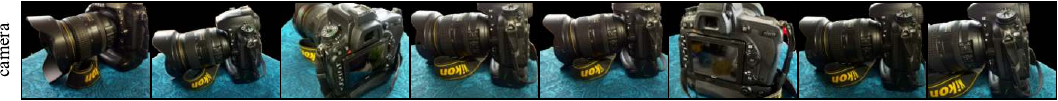}}
    \centerline{\includegraphics[width=\textwidth]{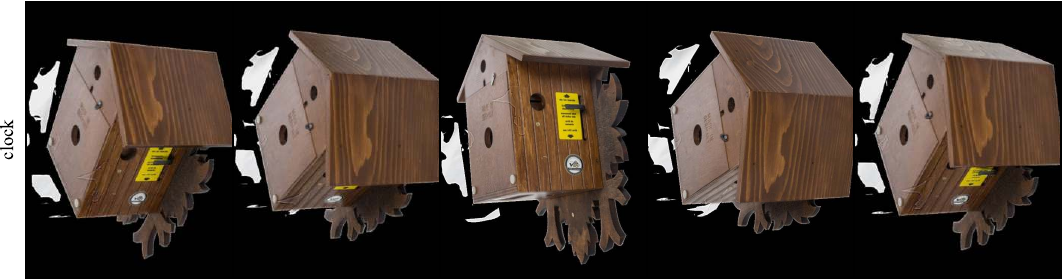}}
    \caption{\textbf{Input images} that we used in our experiments.}
    \label{fig:sup_fig_visible_1}
\end{figure*}

\begin{figure*}[p]
    \centerline{\includegraphics[width=\textwidth]{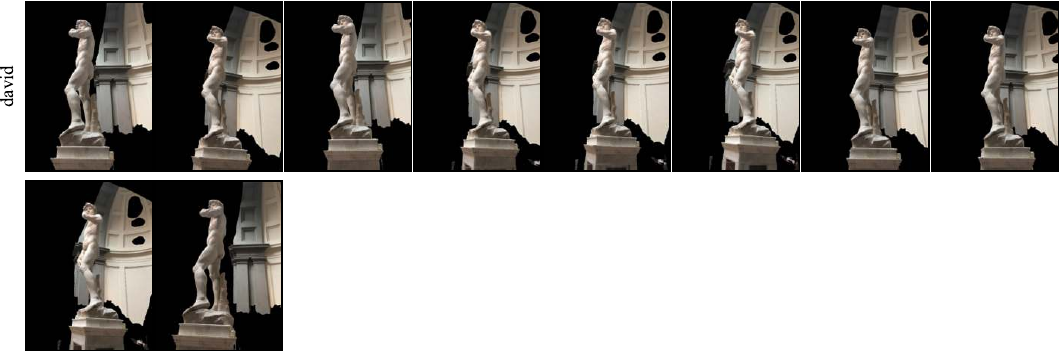}}
    \centerline{\includegraphics[width=\textwidth]{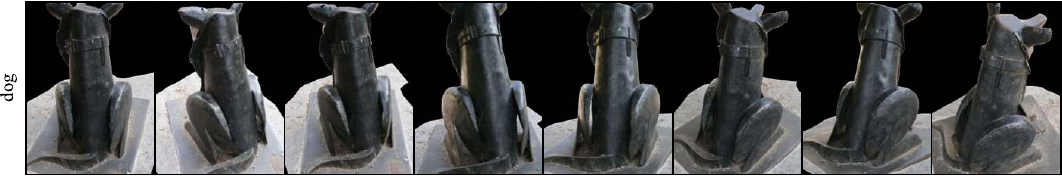}}
    \centerline{\includegraphics[width=\textwidth]{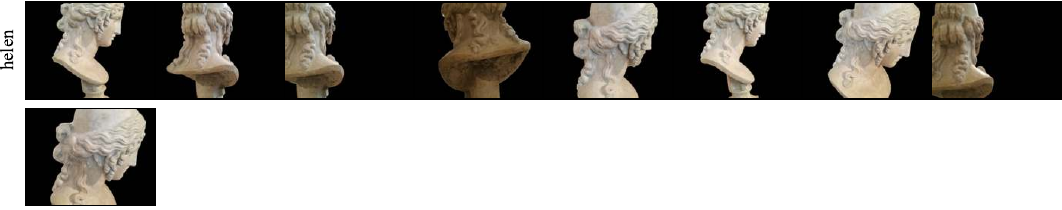}}
    \centerline{\includegraphics[width=\textwidth]{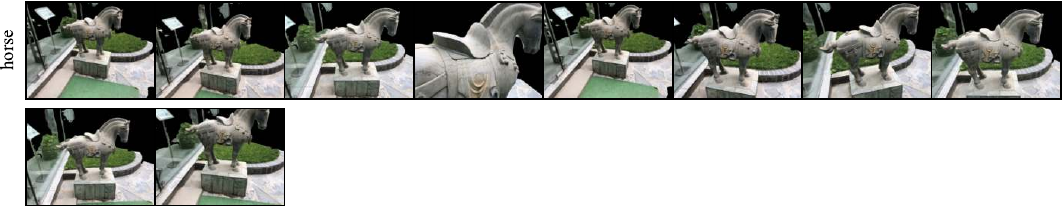}}
    \centerline{\includegraphics[width=\textwidth]{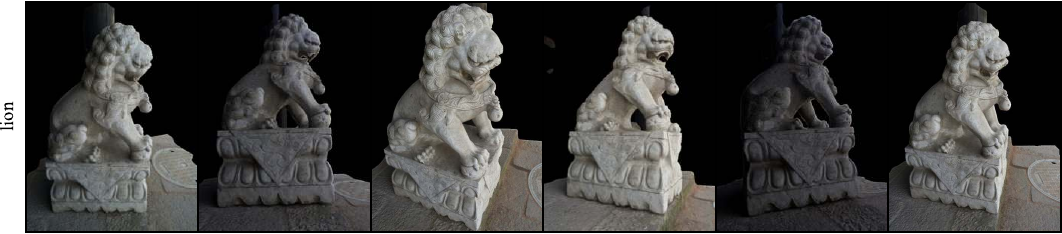}}
    \caption{\textbf{Input images} that we used in our experiments.}
    \label{fig:sup_fig_visible_2}
\end{figure*}

\begin{figure*}[p]
    \centerline{\includegraphics[width=\textwidth]{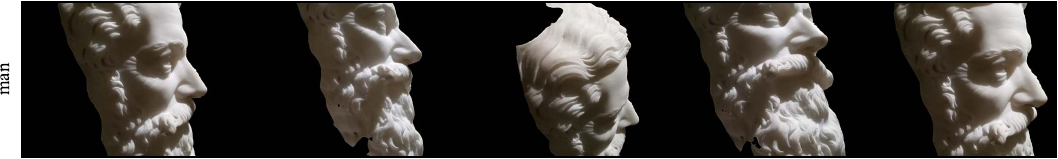}}
    \centerline{\includegraphics[width=\textwidth]{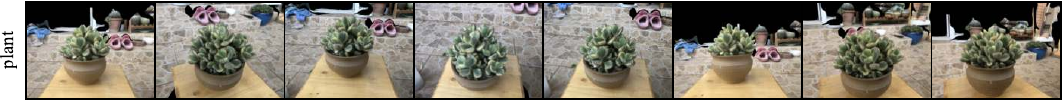}}
    \centerline{\includegraphics[width=\textwidth]{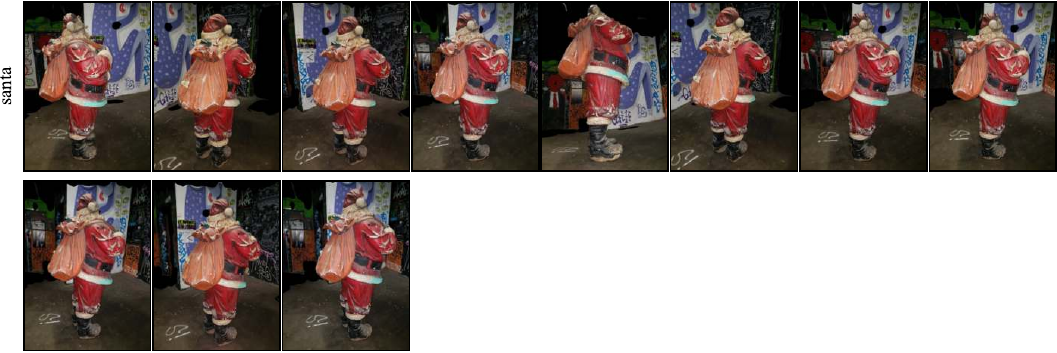}}
    \centerline{\includegraphics[width=\textwidth]{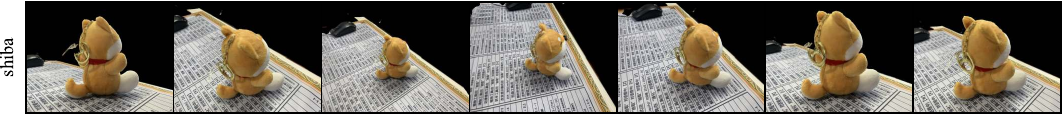}}
    \centerline{\includegraphics[width=\textwidth]{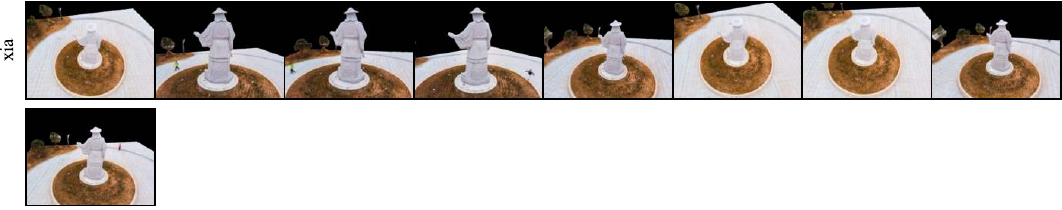}}
    \caption{\textbf{Input images} that we used in our experiments.}
    \label{fig:sup_fig_visible_3}
\end{figure*}

\begin{figure*}[p]
    \centering
    \begin{tikzpicture}
        \node[anchor=south west,inner sep=0] (image) at (0,0) {\includegraphics[width=\textwidth]{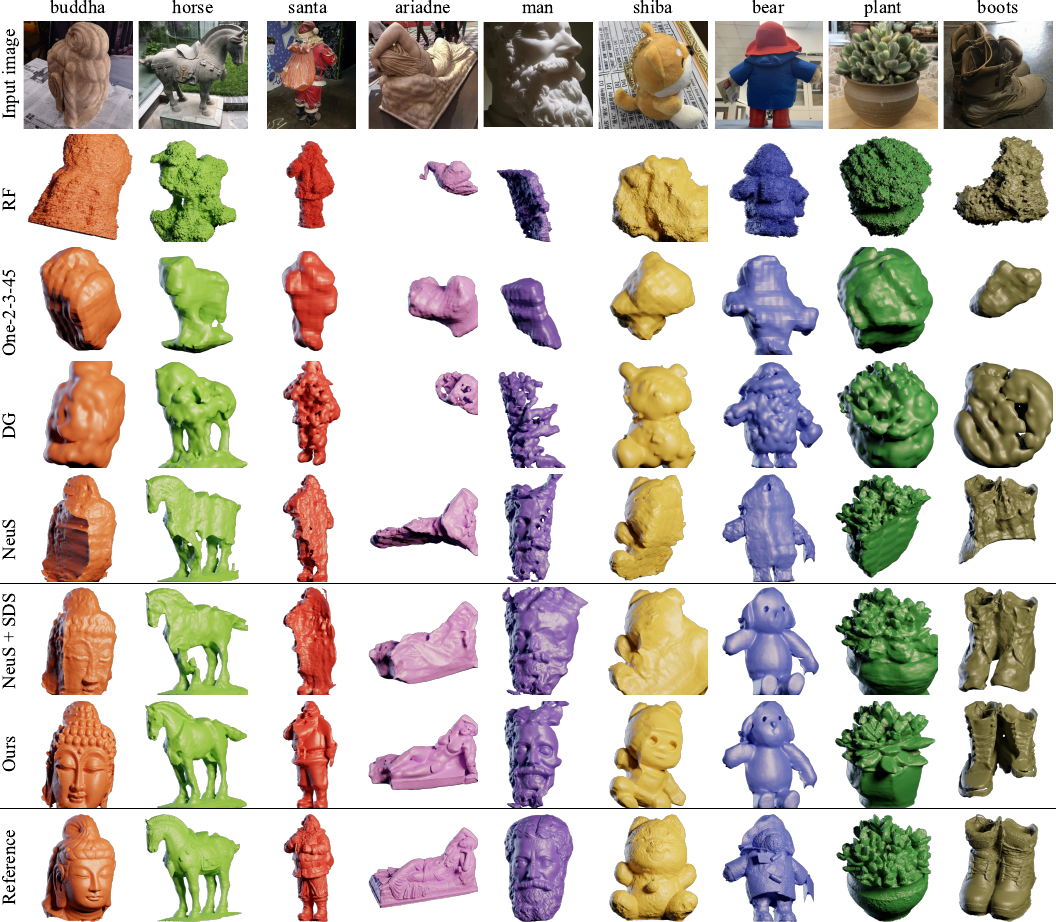}};
        \begin{scope}[shift=(image.north west), x=1pt, y=-1pt, anchor=south west, inner sep=0, font=\small]
            \node at (1.3, 94) {\rotatebox{90}{\phantom{F}~\cite{melas2023realfusion}}};
            \node at (1.3, 131.5) {\rotatebox{90}{\phantom{5}~\cite{liu2023one}}};
            \node at (1.3, 201) {\rotatebox{90}{\phantom{G}~\cite{tang2023dreamgaussian}}};
            \node at (1.3, 249.5) {\rotatebox{90}{\phantom{S}~\cite{wang2021neus}}};
        \end{scope}
    \end{tikzpicture}
    \caption{
        \textbf{Qualitative comparison, unobserved side.}
        The first row shows the image from the visible set used by the competing single-view methods as input. The next four rows show the results of the three generative single-view competitors: RealFusion (RF), One-2-3-45, and DreamGaussian (DG), and the reconstructive competitor (NeuS). The next two rows show the results of the multi-view baseline (NeuS + SDS) and of our full method. The last row depicts the reference surface. All meshes are rendered from the unobserved side.
    }
    \label{fig:sup_fig_qualitative_other}
\end{figure*}

\begin{figure*}[p]
    \centering
    \begin{tikzpicture}
        \node[anchor=south west,inner sep=0] (image) at (0,0) {\includegraphics[width=\textwidth]{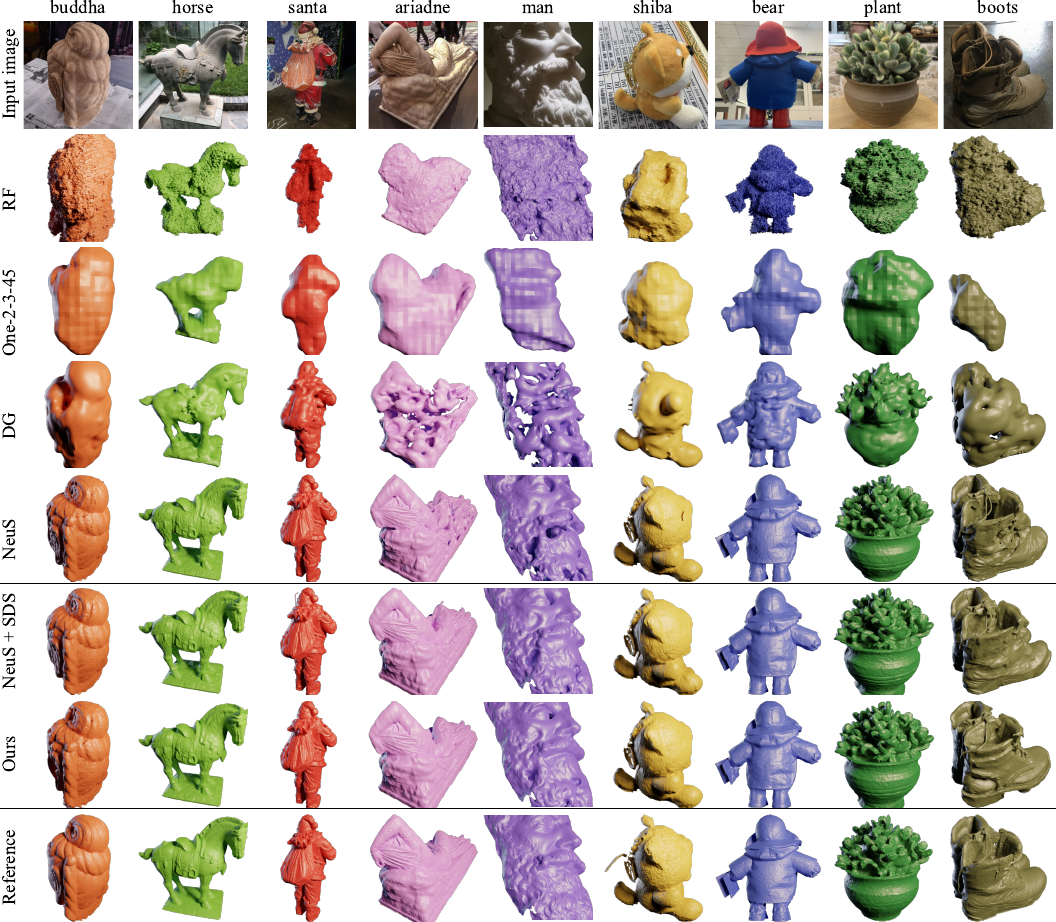}};
        \begin{scope}[shift=(image.north west), x=1pt, y=-1pt, anchor=south west, inner sep=0, font=\small]
            \node at (1.3, 94) {\rotatebox{90}{\phantom{F}~\cite{melas2023realfusion}}};
            \node at (1.3, 131.5) {\rotatebox{90}{\phantom{5}~\cite{liu2023one}}};
            \node at (1.3, 201) {\rotatebox{90}{\phantom{G}~\cite{tang2023dreamgaussian}}};
            \node at (1.3, 249.5) {\rotatebox{90}{\phantom{S}~\cite{wang2021neus}}};
        \end{scope}
    \end{tikzpicture}
    \caption{
        \textbf{Qualitative comparison, visible side.}
        The first row shows the image from the visible set used by the competing single-view methods as input. The next four rows show the results of the three generative single-view competitors: RealFusion (RF), One-2-3-45, and DreamGaussian (DG), and the reconstructive competitor (NeuS). The next two rows show the results of the multi-view baseline (NeuS + SDS) and of our full method. The last row depicts the reference surface. All meshes are rendered from the visible side.
    }
    \label{fig:sup_fig_qualitative_vis_other}
\end{figure*}

\begin{figure*}[p]
    \centering
    \begin{tikzpicture}
        \node[anchor=south west,inner sep=0] (image) at (0,0) {\includegraphics[width=\textwidth]{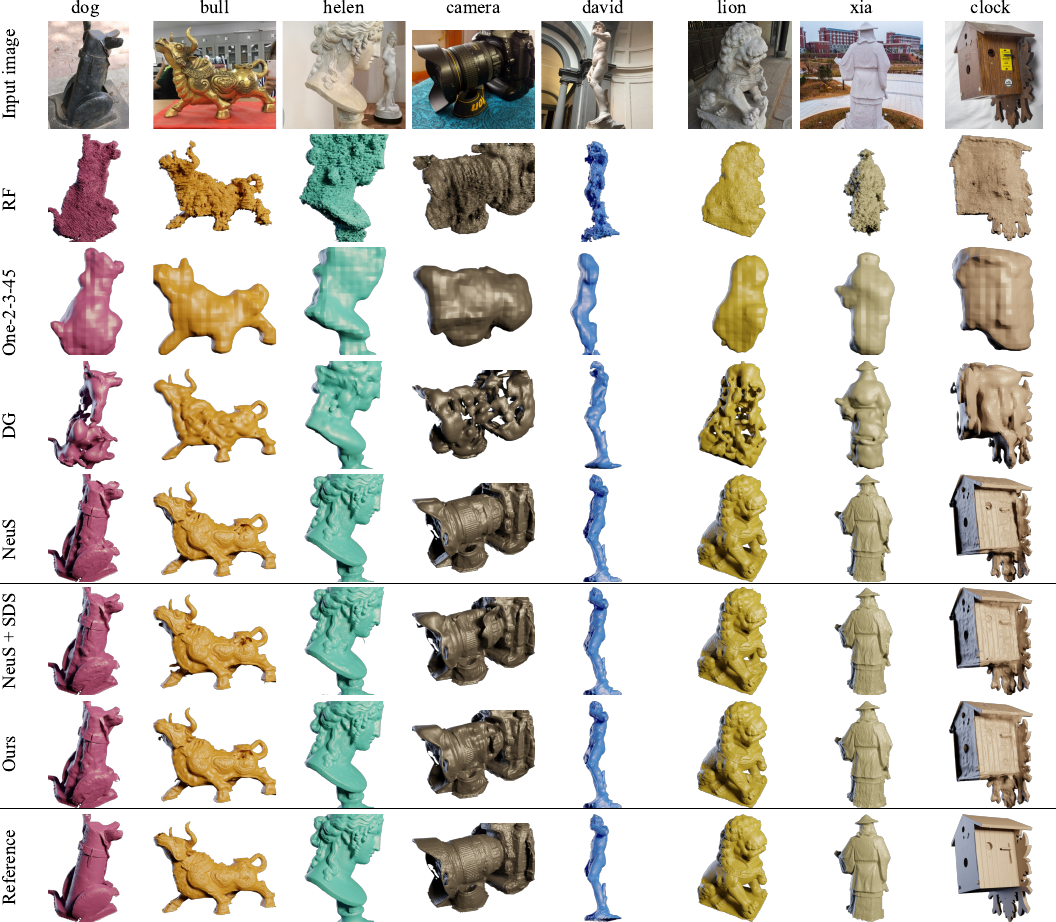}};
        \begin{scope}[shift=(image.north west), x=1pt, y=-1pt, anchor=south west, inner sep=0, font=\small]
            \node at (1.3, 94) {\rotatebox{90}{\phantom{F}~\cite{melas2023realfusion}}};
            \node at (1.3, 131.5) {\rotatebox{90}{\phantom{5}~\cite{liu2023one}}};
            \node at (1.3, 201) {\rotatebox{90}{\phantom{G}~\cite{tang2023dreamgaussian}}};
            \node at (1.3, 249.5) {\rotatebox{90}{\phantom{S}~\cite{wang2021neus}}};
        \end{scope}
    \end{tikzpicture}
    \caption{
        \textbf{Qualitative comparison, visible side.}
        The first row shows the image from the visible set used by the competing single-view methods as input. The next four rows show the results of the three generative single-view competitors: RealFusion (RF), One-2-3-45, and DreamGaussian (DG), and the reconstructive competitor (NeuS). The next two rows show the results of the multi-view baseline (NeuS + SDS) and of our full method. The last row depicts the reference surface. All meshes are rendered from the visible side.
    }
    \label{fig:sup_fig_qualitative_vis_main}
\end{figure*}

\begin{figure*}[p]
    \centerline{\includegraphics[width=\textwidth]{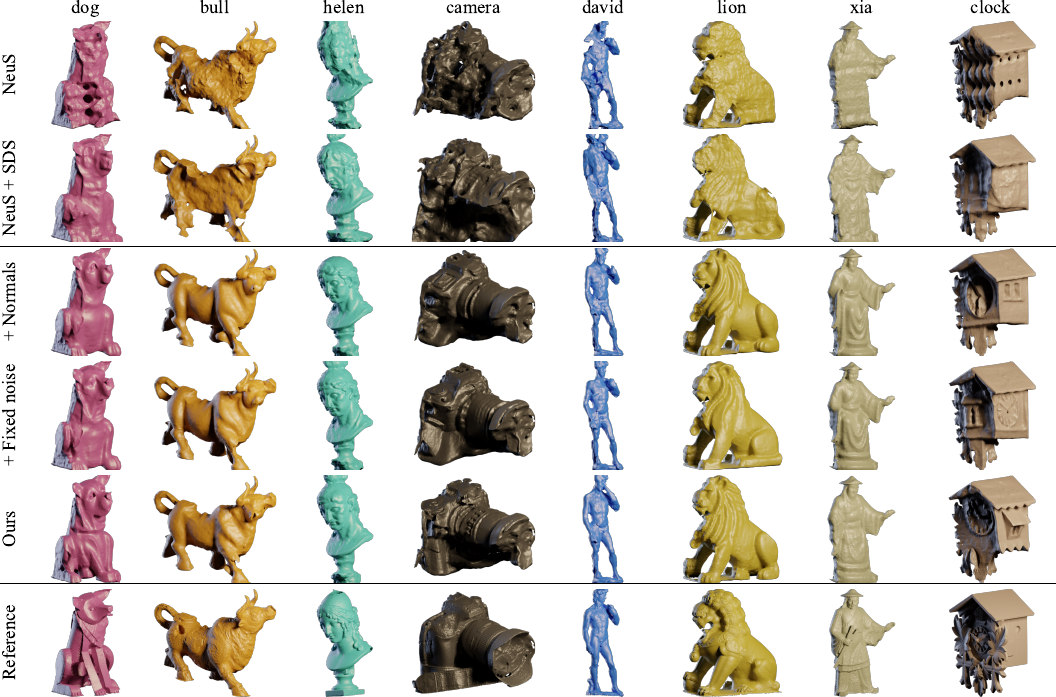}}
    \caption{
        \textbf{Qualitative ablation study.}
        The first two rows show the results of NeuS and our multi-view baseline: NeuS with added color image-based SDS guidance.
        The next three rows show the results produced by successively adding normal maps, frozen SDS, and multi-view SDS guidance.
        The last row depicts the reference surface. All meshes are rendered from the unobserved side.
    }
    \label{fig:sup_fig_ablation_main}
\end{figure*}

\begin{figure*}[p]
    \centerline{\includegraphics[width=\textwidth]{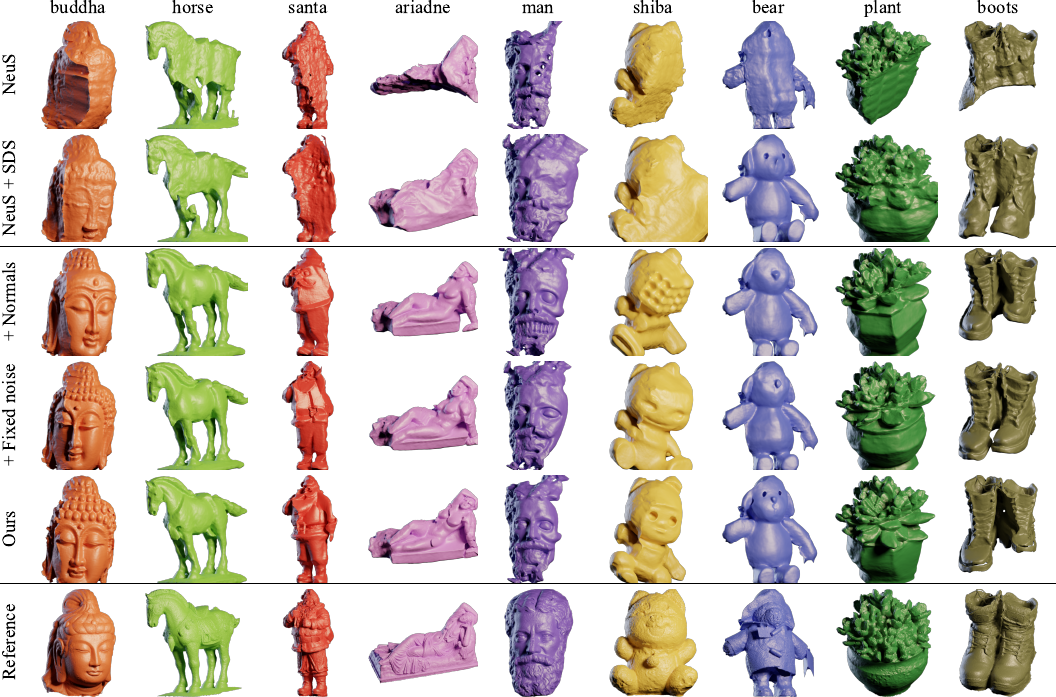}}
    \caption{
        \textbf{Qualitative ablation study.}
        The first two rows show the results of NeuS and our multi-view baseline: NeuS with added color image-based SDS guidance.
        The next three rows show the results produced by successively adding normal maps, frozen SDS, and multi-view SDS guidance.
        The last row depicts the reference surface. All meshes are rendered from the unobserved side.
    }
    \label{fig:sup_fig_ablation_other}
\end{figure*}

We obtain the SDS guidance signal for our method using an open-source diffusion model Stable Diffusion 2.1.
We approach the problem of multi-view inconsistency of the SDS guidance via our multi-view SDS, that we describe in the main text.
In~\cref{fig:sup_fig_multi_view_samples} we show that Stable Diffusion,
conditioned on the prompt \textquote{renders of the same \textlangle object\textrangle{} from different viewpoints},
generates multi-view images of the object composed into a grid with high semantic and stylistic consistency.
This demonstrates that Stable Diffusion can model complex interactions between different viewpoints, if they are combined in this way,
and motivates our approach.

One may notice that the different views of the objects in~\cref{fig:sup_fig_multi_view_samples} may be inconsistent \emph{geometrically}.
We note that in our method we do not use such images directly, but apply SDS guidance to normal maps rendered from the 3D representation of the scene,
that are geometrically multi-view consistent by design.

\section{Theoretical aspects of frozen SDS}
\label{sec:sup_frozen_sds}
The derivation of the Score Distillation Sampling~\cite{poole2022dreamfusion} algorithm starts from the denoising diffusion training loss~\ref{eq:supp_denoising_loss}:
 \begin{align}\label{eq:supp_denoising_loss}   
 \begin{split}
     \mathcal{L}_{Diff}(\phi, \bm{\mathrm{x}}) = \\
     =\mathbb{E}_{t \sim \mathcal{U}(0, 1), \epsilon \sim \mathcal{N}(\bm{0}, \bm{\mathrm{I}})} [w(t) \| \epsilon_{\phi} (\alpha_t \bm{\mathrm{x}} + \sigma_t \epsilon; t) - \epsilon \|_2^2]
     \end{split}
 \end{align}

The gradient of the denoising loss w.r.t. the generated datapoint $\mathrm{\mathbf{x}}$ is taken, and by omitting the U-Net Jacobian loss we get~\ref{eq:supp_sds_denoising_loss}:

\begin{equation}
    \label{eq:supp_sds_denoising_loss} 
    \begin{split}
    \nabla \mathcal{L}_{Diff}(\phi, \bm{\mathrm{x}} = g(\theta)) \overset{\Delta}{=} \\
    \overset{\Delta}{=} \mathbb{E}_{t, \epsilon} \left[ w(t)(\hat{\epsilon}(\mathrm{\mathbf{z}}_t; y, t) - \epsilon) \frac{\partial{\mathrm{\mathbf{x}}}}{\partial{\theta}}\right]
    \end{split}
\end{equation}

While this particular solution may appear to be ad-hoc the authors~\cite{poole2022dreamfusion} show that the same gradient~\ref{eq:supp_sds_denoising_loss} is the gradient of the weighted probability density distillation loss \cite{oord2018representation}.
Both lines of reasoning heavily rely on the assumption that $\epsilon \sim \mathcal{N}(0, \mathrm{\mathbf{I}})$. 
Our results show, that fixing $\epsilon$ and getting rid of the assumption that it belongs to the Gaussian distribution changes very little in terms of outcome.
This leads us to the following conclusions:
a) the motivation, that $\nabla \mathcal{L}_{SDS}$ is the gradient of the denoising diffusion loss w.r.t.  $\mathrm{\mathbf{x}}$  is misleading because the denoising loss needs to be averaged across the distribution of $\epsilon \sim \mathcal{N}(0, \mathrm{\mathbf{I}})$
b) the same equation seemingly produces the unbiased estimation of the gradient of the weighted probability density distillation loss even with the fixed $\epsilon$, with the only variable being $t$. 
Our observations lead us to the conclusion that in its nature the score distillation sampling seems to be much closer to the diffusion inference process than it was previously assumed.
The main problem is that $\mathrm{\mathbf{x}} = g(\theta)$ generally may lie out of the training distribution of the denoising U-net $\epsilon_{\phi}$.
Thus we add the randomly picked at the beginning of the training, but \textbf{fixed} during training noise sample $\epsilon$, with the result now belonging to the training distribution $\bm{\mathrm{x}}^{\prime}_t = \alpha_t \bm{\mathrm{x}} + \sigma_t \epsilon$.
We use the \textquote{shifted} $\bm{\mathrm{x}}^{\prime}_t$ to calculate the update direction and than apply the correction 
$\hat{\epsilon}((\bm{\mathrm{x}}^{\prime}_t; y, t)) - \epsilon)$, which mitigates the impact of the \textquote{shift} and decreases the variance~\cite{poole2022dreamfusion}. 
Our intuition is supported by the fact that most recent works that employ the score distillation sampling to obtain impressive 3D generation results~\cite{zhu2023hifa}, use the decreasing schedule for $t$ resembling the denoising diffusion inference. 

\begin{table*}[p]
    \centerline{\includegraphics[width=\textwidth]{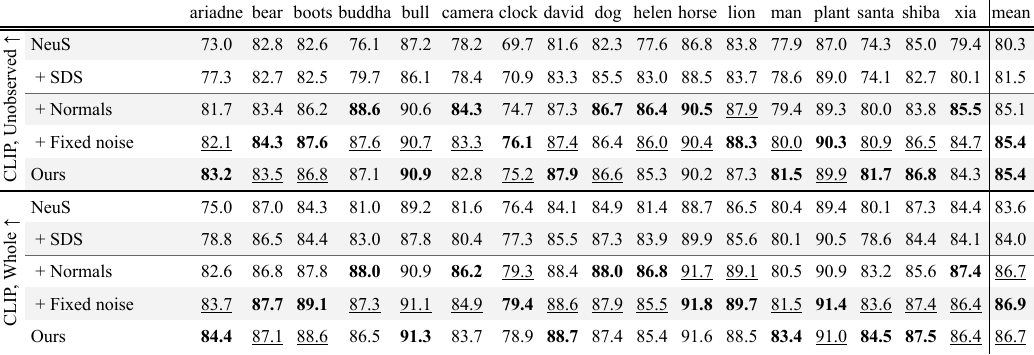}}
    \caption{
    \textbf{Quantitative ablation study.}
    Top: CLIP similarity for the unobserved part of the surface; bottom: for the whole surface.
    The \textbf{best} and \underline{second best} results are highlighted.
    }
    \label{tab:sup_tab_ablation}
\end{table*}

   {\clearpage\newpage\small\bibliographystyle{ieeenat_fullname}\bibliography{src/bib}}

\begin{thebibliography}{62}
\providecommand{\natexlab}[1]{#1}
\providecommand{\url}[1]{\texttt{#1}}
\expandafter\ifx\csname urlstyle\endcsname\relax
  \providecommand{\doi}[1]{doi: #1}\else
  \providecommand{\doi}{doi: \begingroup \urlstyle{rm}\Url}\fi

\bibitem[cli()]{clip_hf}
Clip {ViT-L/14}.
\newblock \url{https://huggingface.co/openai/clip-vit-large-patch14}.

\bibitem[dg_(2023{\natexlab{a}})]{dg_gh_issue1}
A discussion of results of dreamgaussian.
\newblock \url{https://github.com/dreamgaussian/dreamgaussian/issues/15},
  2023{\natexlab{a}}.

\bibitem[dg_(2023{\natexlab{b}})]{dg_gh_issue2}
A discussion of results of dreamgaussian.
\newblock \url{https://github.com/dreamgaussian/dreamgaussian/issues/33},
  2023{\natexlab{b}}.

\bibitem[dg_(2023{\natexlab{c}})]{dg_gh_issue3}
A discussion of results of dreamgaussian.
\newblock \url{https://github.com/dreamgaussian/dreamgaussian/issues/43},
  2023{\natexlab{c}}.

\bibitem[dg_(2023{\natexlab{d}})]{dg_gh_issue4}
A discussion of results of dreamgaussian.
\newblock \url{https://github.com/dreamgaussian/dreamgaussian/issues/65},
  2023{\natexlab{d}}.

\bibitem[mul(2023)]{multiview_dreamgaussian}
Extending dreamgaussian to multi-view setting.
\newblock \url{https://github.com/dreamgaussian/dreamgaussian/issues/22}, 2023.

\bibitem[Aan{\ae}s et~al.(2016)Aan{\ae}s, Jensen, Vogiatzis, Tola, and
  Dahl]{aanaes2016large}
Henrik Aan{\ae}s, Rasmus~Ramsb{\o}l Jensen, George Vogiatzis, Engin Tola, and
  Anders~Bjorholm Dahl.
\newblock Large-scale data for multiple-view stereopsis.
\newblock \emph{International Journal of Computer Vision}, 120:\penalty0
  153--168, 2016.

\bibitem[Armandpour et~al.(2023)Armandpour, Zheng, Sadeghian, Sadeghian, and
  Zhou]{armandpour2023re}
Mohammadreza Armandpour, Huangjie Zheng, Ali Sadeghian, Amir Sadeghian, and
  Mingyuan Zhou.
\newblock Re-imagine the negative prompt algorithm: Transform 2d diffusion into
  3d, alleviate janus problem and beyond.
\newblock \emph{arXiv preprint arXiv:2304.04968}, 2023.

\bibitem[Cai et~al.(2022)Cai, Lin, Zhang, Wang, Wang, and Li]{cai2022learning}
Yingjie Cai, Kwan-Yee Lin, Chao Zhang, Qiang Wang, Xiaogang Wang, and Hongsheng
  Li.
\newblock Learning a structured latent space for unsupervised point cloud
  completion.
\newblock In \emph{Proceedings of the IEEE/CVF Conference on Computer Vision
  and Pattern Recognition}, pages 5543--5553, 2022.

\bibitem[Chang et~al.(2015)Chang, Funkhouser, Guibas, Hanrahan, Huang, Li,
  Savarese, Savva, Song, Su, et~al.]{chang2015shapenet}
Angel~X Chang, Thomas Funkhouser, Leonidas Guibas, Pat Hanrahan, Qixing Huang,
  Zimo Li, Silvio Savarese, Manolis Savva, Shuran Song, Hao Su, et~al.
\newblock Shapenet: An information-rich 3d model repository.
\newblock \emph{arXiv preprint arXiv:1512.03012}, 2015.

\bibitem[Chen et~al.(2023)Chen, Vi{\'e}gas, and Wattenberg]{chen2023beyond}
Yida Chen, Fernanda Vi{\'e}gas, and Martin Wattenberg.
\newblock Beyond surface statistics: Scene representations in a latent
  diffusion model.
\newblock \emph{arXiv preprint arXiv:2306.05720}, 2023.

\bibitem[Cheng et~al.(2023)Cheng, Lee, Tulyakov, Schwing, and
  Gui]{cheng2023sdfusion}
Yen-Chi Cheng, Hsin-Ying Lee, Sergey Tulyakov, Alexander~G Schwing, and
  Liang-Yan Gui.
\newblock Sdfusion: Multimodal 3d shape completion, reconstruction, and
  generation.
\newblock In \emph{Proceedings of the IEEE/CVF Conference on Computer Vision
  and Pattern Recognition}, pages 4456--4465, 2023.

\bibitem[Chu et~al.(2023)Chu, Xie, Mo, Li, Nie{\ss}ner, Fu, and
  Jia]{chu2023diffcomplete}
Ruihang Chu, Enze Xie, Shentong Mo, Zhenguo Li, Matthias Nie{\ss}ner, Chi-Wing
  Fu, and Jiaya Jia.
\newblock Diffcomplete: Diffusion-based generative 3d shape completion.
\newblock \emph{arXiv preprint arXiv:2306.16329}, 2023.

\bibitem[Darmon et~al.(2022)Darmon, Bascle, Devaux, Monasse, and
  Aubry]{darmon2022improving}
Fran{\c{c}}ois Darmon, B{\'e}n{\'e}dicte Bascle, Jean-Cl{\'e}ment Devaux,
  Pascal Monasse, and Mathieu Aubry.
\newblock Improving neural implicit surfaces geometry with patch warping.
\newblock In \emph{Proceedings of the IEEE/CVF Conference on Computer Vision
  and Pattern Recognition}, pages 6260--6269, 2022.

\bibitem[Dogaru et~al.(2023)Dogaru, Ardelean, Ignatyev, Zakharov, and
  Burnaev]{dogaru2023sphere}
Andreea Dogaru, Andrei-Timotei Ardelean, Savva Ignatyev, Egor Zakharov, and
  Evgeny Burnaev.
\newblock Sphere-guided training of neural implicit surfaces.
\newblock In \emph{Proceedings of the IEEE/CVF Conference on Computer Vision
  and Pattern Recognition}, pages 20844--20853, 2023.

\bibitem[Fei et~al.(2022)Fei, Yang, Chen, Li, Li, Ma, Hu, and
  Ma]{fei2022comprehensive}
Ben Fei, Weidong Yang, Wen-Ming Chen, Zhijun Li, Yikang Li, Tao Ma, Xing Hu,
  and Lipeng Ma.
\newblock Comprehensive review of deep learning-based 3d point cloud completion
  processing and analysis.
\newblock \emph{IEEE Transactions on Intelligent Transportation Systems}, 2022.

\bibitem[Fu et~al.(2022)Fu, Xu, Ong, and Tao]{fu2022geo}
Qiancheng Fu, Qingshan Xu, Yew~Soon Ong, and Wenbing Tao.
\newblock Geo-neus: Geometry-consistent neural implicit surfaces learning for
  multi-view reconstruction.
\newblock \emph{Advances in Neural Information Processing Systems},
  35:\penalty0 3403--3416, 2022.

\bibitem[Gal et~al.(2022)Gal, Alaluf, Atzmon, Patashnik, Bermano, Chechik, and
  Cohen-Or]{gal2022image}
Rinon Gal, Yuval Alaluf, Yuval Atzmon, Or Patashnik, Amit~H Bermano, Gal
  Chechik, and Daniel Cohen-Or.
\newblock An image is worth one word: Personalizing text-to-image generation
  using textual inversion.
\newblock \emph{arXiv preprint arXiv:2208.01618}, 2022.

\bibitem[Gropp et~al.(2020)Gropp, Yariv, Haim, Atzmon, and
  Lipman]{gropp2020implicit}
Amos Gropp, Lior Yariv, Niv Haim, Matan Atzmon, and Yaron Lipman.
\newblock Implicit geometric regularization for learning shapes.
\newblock \emph{arXiv preprint arXiv:2002.10099}, 2020.

\bibitem[Jensen et~al.(2014)Jensen, Dahl, Vogiatzis, Tola, and
  Aan{\ae}s]{jensen2014large}
Rasmus Jensen, Anders Dahl, George Vogiatzis, Engin Tola, and Henrik Aan{\ae}s.
\newblock Large scale multi-view stereopsis evaluation.
\newblock In \emph{Proceedings of the IEEE conference on computer vision and
  pattern recognition}, pages 406--413, 2014.

\bibitem[Kasten et~al.(2023)Kasten, Rahamim, and Chechik]{kasten2023point}
Yoni Kasten, Ohad Rahamim, and Gal Chechik.
\newblock Point-cloud completion with pretrained text-to-image diffusion
  models.
\newblock \emph{arXiv preprint arXiv:2306.10533}, 2023.

\bibitem[Kerbl et~al.(2023)Kerbl, Kopanas, Leimk{\"u}hler, and
  Drettakis]{kerbl20233d}
Bernhard Kerbl, Georgios Kopanas, Thomas Leimk{\"u}hler, and George Drettakis.
\newblock 3d gaussian splatting for real-time radiance field rendering.
\newblock \emph{ACM Transactions on Graphics (ToG)}, 42\penalty0 (4):\penalty0
  1--14, 2023.

\bibitem[Knapitsch et~al.(2017)Knapitsch, Park, Zhou, and
  Koltun]{Knapitsch2017}
Arno Knapitsch, Jaesik Park, Qian-Yi Zhou, and Vladlen Koltun.
\newblock Tanks and temples: Benchmarking large-scale scene reconstruction.
\newblock \emph{ACM Transactions on Graphics}, 36\penalty0 (4), 2017.

\bibitem[Lin et~al.(2023)Lin, Gao, Tang, Takikawa, Zeng, Huang, Kreis, Fidler,
  Liu, and Lin]{lin2023magic3d}
Chen-Hsuan Lin, Jun Gao, Luming Tang, Towaki Takikawa, Xiaohui Zeng, Xun Huang,
  Karsten Kreis, Sanja Fidler, Ming-Yu Liu, and Tsung-Yi Lin.
\newblock Magic3d: High-resolution text-to-3d content creation.
\newblock In \emph{Proceedings of the IEEE/CVF Conference on Computer Vision
  and Pattern Recognition}, pages 300--309, 2023.

\bibitem[Liu et~al.(2023{\natexlab{a}})Liu, Xu, Jin, Chen, Xu, Su,
  et~al.]{liu2023one}
Minghua Liu, Chao Xu, Haian Jin, Linghao Chen, Zexiang Xu, Hao Su, et~al.
\newblock One-2-3-45: Any single image to 3d mesh in 45 seconds without
  per-shape optimization.
\newblock \emph{arXiv preprint arXiv:2306.16928}, 2023{\natexlab{a}}.

\bibitem[Liu et~al.(2023{\natexlab{b}})Liu, Wu, Van~Hoorick, Tokmakov,
  Zakharov, and Vondrick]{Liu_2023_ICCV}
Ruoshi Liu, Rundi Wu, Basile Van~Hoorick, Pavel Tokmakov, Sergey Zakharov, and
  Carl Vondrick.
\newblock Zero-1-to-3: Zero-shot one image to 3d object.
\newblock In \emph{Proceedings of the IEEE/CVF International Conference on
  Computer Vision (ICCV)}, pages 9298--9309, 2023{\natexlab{b}}.

\bibitem[Long et~al.(2022)Long, Lin, Wang, Komura, and
  Wang]{long2022sparseneus}
Xiaoxiao Long, Cheng Lin, Peng Wang, Taku Komura, and Wenping Wang.
\newblock Sparseneus: Fast generalizable neural surface reconstruction from
  sparse views.
\newblock In \emph{European Conference on Computer Vision}, pages 210--227.
  Springer, 2022.

\bibitem[Melas-Kyriazi et~al.(2023)Melas-Kyriazi, Laina, Rupprecht, and
  Vedaldi]{melas2023realfusion}
Luke Melas-Kyriazi, Iro Laina, Christian Rupprecht, and Andrea Vedaldi.
\newblock Realfusion: 360deg reconstruction of any object from a single image.
\newblock In \emph{Proceedings of the IEEE/CVF Conference on Computer Vision
  and Pattern Recognition}, pages 8446--8455, 2023.

\bibitem[Mildenhall et~al.(2021)Mildenhall, Srinivasan, Tancik, Barron,
  Ramamoorthi, and Ng]{mildenhall2021nerf}
Ben Mildenhall, Pratul~P Srinivasan, Matthew Tancik, Jonathan~T Barron, Ravi
  Ramamoorthi, and Ren Ng.
\newblock Nerf: Representing scenes as neural radiance fields for view
  synthesis.
\newblock \emph{Communications of the ACM}, 65\penalty0 (1):\penalty0 99--106,
  2021.

\bibitem[M{\"u}ller et~al.(2022)M{\"u}ller, Evans, Schied, and
  Keller]{muller2022instant}
Thomas M{\"u}ller, Alex Evans, Christoph Schied, and Alexander Keller.
\newblock Instant neural graphics primitives with a multiresolution hash
  encoding.
\newblock \emph{ACM Transactions on Graphics (ToG)}, 41\penalty0 (4):\penalty0
  1--15, 2022.

\bibitem[Oord et~al.(2018)Oord, Li, and Vinyals]{oord2018representation}
Aaron van~den Oord, Yazhe Li, and Oriol Vinyals.
\newblock Representation learning with contrastive predictive coding.
\newblock \emph{arXiv preprint arXiv:1807.03748}, 2018.

\bibitem[Poole et~al.(2022)Poole, Jain, Barron, and
  Mildenhall]{poole2022dreamfusion}
Ben Poole, Ajay Jain, Jonathan~T. Barron, and Ben Mildenhall.
\newblock Dreamfusion: Text-to-3d using 2d diffusion.
\newblock \emph{arXiv}, 2022.

\bibitem[Radford et~al.(2021)Radford, Kim, Hallacy, Ramesh, Goh, Agarwal,
  Sastry, Askell, Mishkin, Clark, et~al.]{radford2021learning}
Alec Radford, Jong~Wook Kim, Chris Hallacy, Aditya Ramesh, Gabriel Goh,
  Sandhini Agarwal, Girish Sastry, Amanda Askell, Pamela Mishkin, Jack Clark,
  et~al.
\newblock Learning transferable visual models from natural language
  supervision.
\newblock In \emph{International conference on machine learning}, pages
  8748--8763. PMLR, 2021.

\bibitem[Rombach and Esser()]{stablediffusion_url}
Robin Rombach and Patrick Esser.
\newblock Stable diffusion v2-1.
\newblock \url{https://huggingface.co/stabilityai/stable-diffusion-2-1}.

\bibitem[Rombach et~al.(2022)Rombach, Blattmann, Lorenz, Esser, and
  Ommer]{rombach2022high}
Robin Rombach, Andreas Blattmann, Dominik Lorenz, Patrick Esser, and Bj{\"o}rn
  Ommer.
\newblock High-resolution image synthesis with latent diffusion models.
\newblock In \emph{Proceedings of the IEEE/CVF conference on computer vision
  and pattern recognition}, pages 10684--10695, 2022.

\bibitem[Sch\"{o}nberger and Frahm(2016)]{schoenberger2016structure}
Johannes~Lutz Sch\"{o}nberger and Jan-Michael Frahm.
\newblock {Structure-from-Motion Revisited}.
\newblock In \emph{Conference on Computer Vision and Pattern Recognition
  (CVPR)}, 2016.

\bibitem[Sch\"{o}nberger et~al.(2016)Sch\"{o}nberger, Zheng, Pollefeys, and
  Frahm]{schoenberger2016pixelwise}
Johannes~Lutz Sch\"{o}nberger, Enliang Zheng, Marc Pollefeys, and Jan-Michael
  Frahm.
\newblock {Pixelwise View Selection for Unstructured Multi-View Stereo}.
\newblock In \emph{European Conference on Computer Vision (ECCV)}, 2016.

\bibitem[Sch\"ops et~al.(2017)Sch\"ops, Sch\"onberger, Galliani, Sattler,
  Schindler, Pollefeys, and Geiger]{schoeps2017cvpr}
Thomas Sch\"ops, Johannes~L. Sch\"onberger, Silvano Galliani, Torsten Sattler,
  Konrad Schindler, Marc Pollefeys, and Andreas Geiger.
\newblock A multi-view stereo benchmark with high-resolution images and
  multi-camera videos.
\newblock In \emph{Conference on Computer Vision and Pattern Recognition
  (CVPR)}, 2017.

\bibitem[Tang et~al.(2023)Tang, Ren, Zhou, Liu, and
  Zeng]{tang2023dreamgaussian}
Jiaxiang Tang, Jiawei Ren, Hang Zhou, Ziwei Liu, and Gang Zeng.
\newblock Dreamgaussian: Generative gaussian splatting for efficient 3d content
  creation.
\newblock \emph{arXiv preprint arXiv:2309.16653}, 2023.

\bibitem[Voynov et~al.(2023)Voynov, Bobrovskikh, Karpyshev, Galochkin,
  Ardelean, Bozhenko, Karmanova, Kopanev, Labutin-Rymsho, Rakhimov, Safin,
  Serpiva, Artemov, Burnaev, Tsetserukou, and Zorin]{voynov2023multi}
Oleg Voynov, Gleb Bobrovskikh, Pavel Karpyshev, Saveliy Galochkin,
  Andrei-Timotei Ardelean, Arseniy Bozhenko, Ekaterina Karmanova, Pavel
  Kopanev, Yaroslav Labutin-Rymsho, Ruslan Rakhimov, Aleksandr Safin, Valerii
  Serpiva, Alexey Artemov, Evgeny Burnaev, Dzmitry Tsetserukou, and Denis
  Zorin.
\newblock {Multi-Sensor} {Large-Scale} dataset for {Multi-View} {3D}
  reconstruction.
\newblock In \emph{2023 {IEEE/CVF} Conference on Computer Vision and Pattern
  Recognition ({CVPR})}, pages 21392--21403. IEEE, 2023.

\bibitem[Wang et~al.(2023{\natexlab{a}})Wang, Du, Li, Yeh, and
  Shakhnarovich]{wang2023score}
Haochen Wang, Xiaodan Du, Jiahao Li, Raymond~A Yeh, and Greg Shakhnarovich.
\newblock Score jacobian chaining: Lifting pretrained 2d diffusion models for
  3d generation.
\newblock In \emph{Proceedings of the IEEE/CVF Conference on Computer Vision
  and Pattern Recognition}, pages 12619--12629, 2023{\natexlab{a}}.

\bibitem[Wang et~al.(2022{\natexlab{a}})Wang, Wang, Long, Theobalt, Komura,
  Liu, and Wang]{wang2022neuris}
Jiepeng Wang, Peng Wang, Xiaoxiao Long, Christian Theobalt, Taku Komura,
  Lingjie Liu, and Wenping Wang.
\newblock Neuris: Neural reconstruction of indoor scenes using normal priors.
\newblock In \emph{European Conference on Computer Vision}, pages 139--155.
  Springer, 2022{\natexlab{a}}.

\bibitem[Wang et~al.(2021)Wang, Liu, Liu, Theobalt, Komura, and
  Wang]{wang2021neus}
Peng Wang, Lingjie Liu, Yuan Liu, Christian Theobalt, Taku Komura, and Wenping
  Wang.
\newblock Neus: Learning neural implicit surfaces by volume rendering for
  multi-view reconstruction.
\newblock \emph{NeurIPS}, 2021.

\bibitem[Wang et~al.(2022{\natexlab{b}})Wang, Li, Jiang, Zhou, Cao, Fu, and
  Xiao]{wang2022neuralroom}
Yusen Wang, Zongcheng Li, Yu Jiang, Kaixuan Zhou, Tuo Cao, Yanping Fu, and
  Chunxia Xiao.
\newblock Neuralroom: Geometry-constrained neural implicit surfaces for indoor
  scene reconstruction.
\newblock \emph{arXiv preprint arXiv:2210.06853}, 2022{\natexlab{b}}.

\bibitem[Wang et~al.(2023{\natexlab{b}})Wang, Han, Habermann, Daniilidis,
  Theobalt, and Liu]{wang2023neus2}
Yiming Wang, Qin Han, Marc Habermann, Kostas Daniilidis, Christian Theobalt,
  and Lingjie Liu.
\newblock Neus2: Fast learning of neural implicit surfaces for multi-view
  reconstruction.
\newblock In \emph{Proceedings of the IEEE/CVF International Conference on
  Computer Vision}, pages 3295--3306, 2023{\natexlab{b}}.

\bibitem[Wang et~al.(2004)Wang, Bovik, Sheikh, and Simoncelli]{wang2004image}
Zhou Wang, Alan~C Bovik, Hamid~R Sheikh, and Eero~P Simoncelli.
\newblock Image quality assessment: from error visibility to structural
  similarity.
\newblock \emph{IEEE transactions on image processing}, 13\penalty0
  (4):\penalty0 600--612, 2004.

\bibitem[Wu et~al.(2020)Wu, Chen, Zhuang, and Chen]{wu2020multimodal}
Rundi Wu, Xuelin Chen, Yixin Zhuang, and Baoquan Chen.
\newblock Multimodal shape completion via conditional generative adversarial
  networks.
\newblock In \emph{Computer Vision--ECCV 2020: 16th European Conference,
  Glasgow, UK, August 23--28, 2020, Proceedings, Part IV 16}, pages 281--296.
  Springer, 2020.

\bibitem[Xiang et~al.(2021)Xiang, Wen, Liu, Cao, Wan, Zheng, and
  Han]{xiang2021snowflakenet}
Peng Xiang, Xin Wen, Yu-Shen Liu, Yan-Pei Cao, Pengfei Wan, Wen Zheng, and
  Zhizhong Han.
\newblock Snowflakenet: Point cloud completion by snowflake point deconvolution
  with skip-transformer.
\newblock In \emph{Proceedings of the IEEE/CVF international conference on
  computer vision}, pages 5499--5509, 2021.

\bibitem[Yan et~al.(2022)Yan, Lin, Mitra, Lischinski, Cohen-Or, and
  Huang]{yan2022shapeformer}
Xingguang Yan, Liqiang Lin, Niloy~J Mitra, Dani Lischinski, Daniel Cohen-Or,
  and Hui Huang.
\newblock Shapeformer: Transformer-based shape completion via sparse
  representation.
\newblock In \emph{Proceedings of the IEEE/CVF Conference on Computer Vision
  and Pattern Recognition}, pages 6239--6249, 2022.

\bibitem[Yao et~al.(2018)Yao, Luo, Li, Fang, and Quan]{yao2018mvsnet}
Yao Yao, Zixin Luo, Shiwei Li, Tian Fang, and Long Quan.
\newblock Mvsnet: Depth inference for unstructured multi-view stereo.
\newblock \emph{European Conference on Computer Vision (ECCV)}, 2018.

\bibitem[Yao et~al.(2020)Yao, Luo, Li, Zhang, Ren, Zhou, Fang, and
  Quan]{yao2020blendedmvs}
Yao Yao, Zixin Luo, Shiwei Li, Jingyang Zhang, Yufan Ren, Lei Zhou, Tian Fang,
  and Long Quan.
\newblock Blendedmvs: A large-scale dataset for generalized multi-view stereo
  networks.
\newblock In \emph{Proceedings of the IEEE/CVF conference on computer vision
  and pattern recognition}, pages 1790--1799, 2020.

\bibitem[Yariv et~al.(2020)Yariv, Kasten, Moran, Galun, Atzmon, Ronen, and
  Lipman]{yariv2020multiview}
Lior Yariv, Yoni Kasten, Dror Moran, Meirav Galun, Matan Atzmon, Basri Ronen,
  and Yaron Lipman.
\newblock Multiview neural surface reconstruction by disentangling geometry and
  appearance.
\newblock \emph{Advances in Neural Information Processing Systems},
  33:\penalty0 2492--2502, 2020.

\bibitem[Yariv et~al.(2021)Yariv, Gu, Kasten, and Lipman]{yariv2021volume}
Lior Yariv, Jiatao Gu, Yoni Kasten, and Yaron Lipman.
\newblock Volume rendering of neural implicit surfaces.
\newblock \emph{Advances in Neural Information Processing Systems},
  34:\penalty0 4805--4815, 2021.

\bibitem[Yu et~al.(2021{\natexlab{a}})Yu, Ye, Tancik, and
  Kanazawa]{yu2020pixelnerf}
Alex Yu, Vickie Ye, Matthew Tancik, and Angjoo Kanazawa.
\newblock {pixelNeRF}: Neural radiance fields from one or few images.
\newblock In \emph{CVPR}, 2021{\natexlab{a}}.

\bibitem[Yu et~al.(2021{\natexlab{b}})Yu, Rao, Wang, Liu, Lu, and
  Zhou]{yu2021pointr}
Xumin Yu, Yongming Rao, Ziyi Wang, Zuyan Liu, Jiwen Lu, and Jie Zhou.
\newblock Pointr: Diverse point cloud completion with geometry-aware
  transformers.
\newblock In \emph{Proceedings of the IEEE/CVF international conference on
  computer vision}, pages 12498--12507, 2021{\natexlab{b}}.

\bibitem[Zacharov et~al.(2019)Zacharov, Arslanov, Gunin, Stefonishin, Bykov,
  Pavlov, Panarin, Maliutin, Rykovanov, and Fedorov]{zacharov2019zhores}
Igor Zacharov, Rinat Arslanov, Maksim Gunin, Daniil Stefonishin, Andrey Bykov,
  Sergey Pavlov, Oleg Panarin, Anton Maliutin, Sergey Rykovanov, and Maxim
  Fedorov.
\newblock ``zhores''-petaflops supercomputer for data-driven modeling, machine
  learning and artificial intelligence installed in skolkovo institute of
  science and technology.
\newblock \emph{Open Engineering}, 9\penalty0 (1):\penalty0 512--520, 2019.

\bibitem[Zhang et~al.(2021)Zhang, Chen, Cai, Pan, Zhao, Yi, Yeo, Dai, and
  Loy]{zhang2021unsupervised}
Junzhe Zhang, Xinyi Chen, Zhongang Cai, Liang Pan, Haiyu Zhao, Shuai Yi,
  Chai~Kiat Yeo, Bo Dai, and Chen~Change Loy.
\newblock Unsupervised 3d shape completion through gan inversion.
\newblock In \emph{Proceedings of the IEEE/CVF Conference on Computer Vision
  and Pattern Recognition}, pages 1768--1777, 2021.

\bibitem[Zhang et~al.(2023)Zhang, Rao, and Agrawala]{zhang2023adding}
Lvmin Zhang, Anyi Rao, and Maneesh Agrawala.
\newblock Adding conditional control to text-to-image diffusion models.
\newblock In \emph{Proceedings of the IEEE/CVF International Conference on
  Computer Vision}, pages 3836--3847, 2023.

\bibitem[Zhang et~al.(2018)Zhang, Isola, Efros, Shechtman, and
  Wang]{zhang2018perceptual}
Richard Zhang, Phillip Isola, Alexei~A Efros, Eli Shechtman, and Oliver Wang.
\newblock The unreasonable effectiveness of deep features as a perceptual
  metric.
\newblock In \emph{CVPR}, 2018.

\bibitem[Zhou et~al.(2021)Zhou, Du, and Wu]{zhou20213d}
Linqi Zhou, Yilun Du, and Jiajun Wu.
\newblock 3d shape generation and completion through point-voxel diffusion.
\newblock In \emph{Proceedings of the IEEE/CVF International Conference on
  Computer Vision}, pages 5826--5835, 2021.

\bibitem[Zhou and Tulsiani(2023)]{zhou2023sparsefusion}
Zhizhuo Zhou and Shubham Tulsiani.
\newblock Sparsefusion: Distilling view-conditioned diffusion for 3d
  reconstruction.
\newblock In \emph{Proceedings of the IEEE/CVF Conference on Computer Vision
  and Pattern Recognition}, pages 12588--12597, 2023.

\bibitem[Zhu and Zhuang(2023)]{zhu2023hifa}
Joseph Zhu and Peiye Zhuang.
\newblock Hifa: High-fidelity text-to-3d with advanced diffusion guidance.
\newblock \emph{arXiv preprint arXiv:2305.18766}, 2023.

\end{thebibliography}
\end{document}